\documentclass[runningheads]{llncs}
\usepackage{graphicx}

\usepackage{xcolor}
\usepackage{ifthen}
\usepackage{textcomp}
\usepackage{color}
\usepackage{subfigure}

\usepackage{tikz}
\usepackage{comment}
\usepackage{amsmath,amssymb} % define this before the line numbering.

\usepackage{microtype}
\usepackage{graphicx}
\usepackage{subfigure}
\usepackage{booktabs} % for professional tables
\usepackage{enumitem}
\usepackage{makecell}
\usepackage{multirow}
\usepackage{hyperref}

\usepackage{caption}

\usepackage[accsupp]{axessibility}  % Improves PDF readability for those with disabilities.

\newcommand{\boldstart}[1]{\noindent\textbf{#1}}
\newcommand{\boldstartspace}[1]{\vspace{0.1in}\noindent\textbf{#1}}
\newcommand{\fv}[1]{\textcolor{black}{#1}}

\makeatletter
\newcommand{\printfnsymbol}[1]{%
        \textsuperscript{\@fnsymbol{#1}}%
}
\makeatother

\begin{document}
% \renewcommand\thelinenumber{\color[rgb]{0.2,0.5,0.8}\normalfont\sffamily\scriptsize\arabic{linenumber}\color[rgb]{0,0,0}}
% \renewcommand\makeLineNumber {\hss\thelinenumber\ \hspace{6mm} \rlap{\hskip\textwidth\ \hspace{6.5mm}\thelinenumber}}
% \linenumbers
\pagestyle{headings}
\mainmatter

\title{TensoRF: Tensorial Radiance Fields} % Replace with your title

% CAMERA READY SUBMISSION
% \begin{comment}
\titlerunning{TensoRF: Tensorial Radiance Fields}
% If the paper title is too long for the running head, you can set
% an abbreviated paper title here
%

\author{Anpei Chen$^1$\thanks{Equal contribution.\\ Research done when Anpei Chen was in a remote internship with UCSD.}
% For a paper whose authors are all at the same institution,
% omit the following lines up until the closing ``}''.
% Additional authors and addresses can be added with ``\and'',
% just like the second author.
% To save space, use either the email address or home page, not both
\,\,
Zexiang Xu$^{2}$\printfnsymbol{1}
\,\,
Andreas Geiger$^3$ 
\,\,
Jingyi Yu$^1$
\,\,
Hao Su$^4$ 
}
\institute{$^1$ShanghaiTech University \quad $^2$Adobe Research \\ $^3$University of Tübingen and MPI-IS, Tübingen \quad $^4$UC San Diego
\href{https://apchenstu.github.io/TensoRF/}{https://apchenstu.github.io/TensoRF/}
}

\authorrunning{A. Chen, Z. Xu et al.}
%
% \end{comment}
%******************

\maketitle

\begin{abstract}
We present TensoRF, a novel approach to model and reconstruct radiance fields.
Unlike NeRF that purely uses MLPs, we model the radiance field of a scene as a 4D tensor, which represents a 3D voxel grid with per-voxel multi-channel features.
Our central idea is to factorize the 4D scene tensor into multiple compact low-rank tensor components. 
% We find that we can already get results better than NeRF by applying the classic CP decomposition that factorizes tensors into rank-one components with pure compact vectors.
We demonstrate that applying traditional \fv{CANDECOMP/PARAFAC (CP)} decomposition -- that factorizes tensors into rank-one components with compact vectors -- in our framework leads to improvements over vanilla NeRF. 
% To further boost performance, we further introduce a novel vector-matrix (VM) decomposition that .
To further boost performance, we introduce a novel vector-matrix (VM) decomposition that relaxes the low-rank constraints for two modes of a tensor and factorizes tensors into compact vector and matrix factors. 
Beyond superior rendering quality, % are both of high compactness and support efficient radiance field evaluation. 
our models with CP and VM decompositions lead to a significantly lower memory footprint in comparison to previous and concurrent works that directly optimize per-voxel features.
Experimentally, we demonstrate that TensoRF with CP decomposition achieves fast reconstruction ($<30$ min) with better rendering quality and even a smaller model size ($<4$ MB) compared to NeRF.
Moreover, TensoRF with VM decomposition further boosts rendering quality and outperforms previous state-of-the-art methods, while reducing the reconstruction time ($<10$ min) and retaining a compact model size ($<75$ MB).

% These components can be effectively reconstructed 
% %from captured images 
% with differentiable radiance field rendering and gradient descent.
% Inspired by classical tensor decomposition techniques, we also present a novel tensor decomposition that relaxes the low-rank constraints for two modes of a tensor and factorize tensors into compact vector and matrix factors.
% Our model is of high compactness and supports efficient evaluation of volume rendering properties.
% % This leads to highly efficient radiance field reconstruction and compression.
% %, while the model's compactness is relatively decreased.
% Our tensorial radiance fields lead to photo-realistic novel view synthesis on par with the state-of-the-art results, while enabling fast optimization ($<$30min) and compact modeling ($<$60MB). 

% % Our tensorial radiance fields lead to photo-realistic novel view synthesis on par with the state-of-the-art, while consuming substantially less memory. 
\end{abstract}

\newcommand{\Tensor}{\mathcal{T}}
\newcommand{\Vector}{\mathbf{v}}
\newcommand{\Matrix}{\mathbf{M}}
\newcommand{\AppVec}{\mathbf{b}}
\newcommand{\AppMat}{\mathbf{B}}
\newcommand{\VectorPack}{\mathbf{V}}
\newcommand{\RRR}{\mathbb{R}}
\newcommand{\DimIJK}{I\times J\times K}
\newcommand{\DimCh}{P}

\newcommand{\OuterP}{\circ}
\newcommand{\ScalarP}{\ast}
\newcommand{\Pos}{\mathbf{x}}
\newcommand{\Dens}{\sigma}
\newcommand{\Rad}{c}
\newcommand{\Color}{C}
\newcommand{\Trans}{\tau}
\newcommand{\Step}{\Delta}
\newcommand{\Comp}{\mathcal{A}}
\newcommand{\Dir}{d}
\newcommand{\Grid}{\mathcal{G}}
\newcommand{\ShadFunc}{S}
\newcommand{\loss}{\mathcal{L}}

\begin{figure}[t]
    \includegraphics[width=\linewidth]{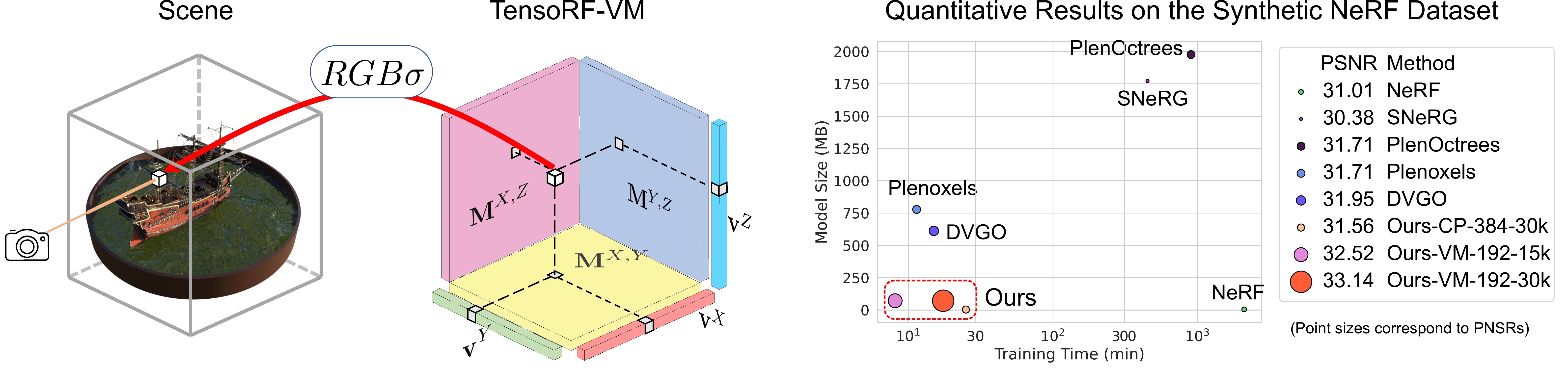}
    \vspace{-5mm}
    \caption{
    Left: We model a scene as a tensorial radiance field using a set of vectors ($\Vector$) and matrices ($\Matrix$) that describe scene appearance and geometry along their corresponding axes. These vector/matrix factors are used to compute volume density $\Dens$ and view-dependent RGB color via vector-matrix outer products for realistic volume rendering.
    Right: In comparison with previous and concurrent methods, our TensoRF models can achieve the best rendering quality and are the only methods that can simultaneously achieve fast reconstruction and high compactness. (Our models are denoted with their decomposition techniques, number of components, and training steps.)
    % We model a scene (a) as a tensorial radiance field (b) using a set of vectors ($\Vector$) and matrices ($\Matrix$) that describe scene appearance and geometry along their corresponding axes. These vector/matrix factors are used to compute volume density $\Dens$ and view-dependent RGB color via vector-matrix outer products, leading to efficient radiance field reconstruction and realistic rendering (c). 
    } \vspace{-5mm}
    \label{fig:teaser}
\end{figure}

\section{Introduction}

Modeling and reconstructing 3D scenes as representations that support high-quality image synthesis is crucial for computer vision and graphics with various applications in visual effects, e-commerce, virtual and augmented reality, and robotics.
Recently, NeRF \cite{mildenhall2020nerf} and its many follow-up works \cite{zhang2020nerf++,liu2020neural} have shown success on modeling a scene as a radiance field 
% -- a continuous field of volume density and view-dependent color for differentiable volume rendering -- 
and enabled photo-realistic rendering of scenes with highly complex geometry and view-dependent appearance effects. Despite the fact that (purely MLP-based) NeRF models require small memory, they take a long time (hours or days) to train. In this work, we pursue a novel approach that is both \emph{efficient in training time} and \emph{compact in memory footprint}, and at the same time achieves \emph{state-of-the-art} rendering quality.
% By modeling scenes as volumetric radiance fields, NeRF and its many following works have shown great success on various image synthesis applications, such as view synthesis, appearance acquisition, and generative models.
% While many methods use pure MLPs to represent radiance fields as the original NeRF, many recent methods adopt a voxel grid of per-voxel features  

% Note that NeRF learns a network to predict radiance at each voxel, but training a deep neural network is not easy. To speed up the optimization, it is natural to set optimization variables to be radiance in space directly. For example, one can model the radiance field as a 3D voxel. 
% For example, several previous or concurrent efforts \cite{liu2020neural,yu2021plenoctrees,hedman2021baking} adopt an explicit 3D voxel grid of per-voxel features in place of the pure coordinate-based MLPs in the original NeRF. % for representing radiance fields, 
% Several concurrent works optimize , while promising, these voxel-based methods are still inefficient on memory usage and reconstruction time: They generally require large GPU memory to store all voxels whose size grows cubically with resolution, and some even require pre-computing an MLP-based NeRF for distillation, leading to very long reconstruction time.

To do so, we propose TensoRF, a novel radiance field representation that is highly compact and also fast to reconstruct, enabling efficient scene reconstruction and modeling.
% While we also consider a 3D voxel grid of features (which can be either neural features or simple spherical harmonics coefficients), our approach advances its efficiency significantly with a novel tensorial approach.
Unlike coordinate-based MLPs used in NeRF, we represent radiance fields as an explicit voxel grid of features.
Note that it is \emph{unclear} whether voxel grid representation can benefit the efficiency of reconstruction: While previous work has used feature grids \cite{liu2020neural,yu2021plenoctrees,hedman2021baking}, they require large GPU memory to store the voxels whose size grows cubically with resolution, and some even require pre-computing a NeRF for distillation, leading to very long reconstruction time.

Our work addresses the inefficiency of voxel grid representations in a principled framework, leading to a family of simple yet effective methods.
% Our approach extends voxel grid-based methods and advances its efficiency significantly.
% We essentially address the costly issue in previous voxel-based radiance fields with a novel tensorial approach.
We leverage the fact that a feature grid can naturally be seen as a 4D tensor, where three of its modes correspond to the XYZ axes of the grid and the fourth mode represents the feature channel dimension.
This opens the possibility of exploiting classical tensor decomposition techniques -- which have been widely applied to high-dimensional data analysis and compression in various fields \cite{kolda2009tensor} --  for radiance field modeling.  % This allows us to combine radiance fields with classical tensor decomposition techniques. 
% Inspired by classical tensor decomposition techniques,  % , which encode the entire information of a voxel feature grid that covers the 3D scene 
%(CANDECOMP/PARAFAC) 
We, therefore, propose to factorize the tensor of radiance fields into multiple \emph{low-rank} tensor components, leading to an accurate and compact scene representation.
Note that our central idea of tensorizing radiance fields is general and can be potentially adopted to any tensor decomposition technique. 

In this work, we first attempt the classic CANDECOMP/PARAFAC (CP) decomposition \cite{carroll1970analysis}. %and also introduce a new vector-matrix (VM) decomposition for representing TensoRFs.
% In this work, we apply the classic CANDECOMP/PARAFAC (CP) decomposition \cite{carroll1970analysis} and also introduce a novel vector-matrix (VM) decomposition for representing TensoRFs.
We show that TensoRF with CP decomposition can already achieve photo-realistic rendering and lead to a more compact model than NeRF that is purely MLP based (see Fig.~\ref{fig:teaser} and Tab.~\ref{table:results}).
However, experimentally, to further push reconstruction quality for complex scenes, we have to use more component factors, which undesirably increases training time. 

Therefore, we present a novel vector-matrix (VM) decomposition technique that effectively reduces the number of components required for the same expression capacity, leading to faster reconstruction and better rendering.
% leading to faster radiance field reconstruction without sacrificing high compactness too much.
% However, classical methods, like CP decomposition \cite{carroll1970analysis},  mainly focus on high compactness with low-rank approximation.
% Hence, a large number of component factors can be required to express the radiance field tensors with high fidelity, causing high computational costs in recovering tensor elements, undesirably slowing down the process of radiance field reconstruction and rendering.
% Therefore, we introduce a novel decomposition technique that enables efficient recovery of tensor elements for radiance field reconstruction 
% without sacrificing high compactness too much.
In particular, inspired by the CP and block term decomposition \cite{de2008decompositions}, we propose to factorize the full tensor of a radiance field into multiple vector and matrix factors per tensor component. 
% In contrast to CP decomposition that purely uses vectors as factors (with vector outer products), 
Unlike the sum of outer products of pure vectors in CP decomposition, we consider the sum of vector-matrix outer products (see Fig.~\ref{fig:tensor-decomp}).
% CP decomposition constrains each mode of a component to be strictly rank-one by expressing it with a single vector factor, 
% whereas we model two modes jointly as a matrix factor for each component.
% In essence, we increase the number of parameters in each component for two modes and relax the ranks of the two modes to be arbitrarily large.
In essence, we relax the ranks of two modes of each component by jointly modeling two modes in a matrix factor.
While this increases the model size compared to pure vector-based factorization in CP, we enable each component to express more complex tensor data of higher ranks, thus significantly reducing the required number of components in radiance field modeling.
% and leading to better rendering quality (see Fig.~\ref{fig:teaser} and Tab.~\ref{tab:decomp}).

% On the other hand, while our decomposition is less compact than CP and other standard low-rank decomposition, our model can still effectively compress the tensor data, leading to much smaller memory footprint than using a dense feature grid.

% In essence, we represent a radiance field of a scene by a set of compact factors of tensor components. 
% Without necessarily recovering the full feature grid, these component factors can decode a single feature vector at any given 3D location for computing volume density and view-dependent color, effectively expressing a radiance field and supporting volumetric radiance field rendering as in NeRF.

With CP/VM decomposition, our approach compactly encodes spatially varying features in the voxel grid. Volume density and view-dependent color can be decoded from the features, supporting volumetric radiance field rendering. Because a tensor expresses discrete data, we also enable efficient trilinear interpolation for our representation to model a continuous field.
Our representation supports various types of per-voxel features with different decoding functions, %used in previous voxel-based methods
including neural features -- depending on an MLP to regress view-dependent colors from the features -- and spherical harmonics (SH) features (coefficients) -- allowing for simple color computation from the fixed SH functions and leading to a representation without neural networks.

Our tensorial radiance fields can be effectively reconstructed from multi-view images and enable realistic novel view synthesis.
In contrast to previous works that directly reconstruct voxels, our tensor factorization reduces space complexity from $\mathcal{O}(n^3)$ to $\mathcal{O}(n)$ (with CP) or $\mathcal{O}(n^2)$ (with VM), significantly lowering memory footprint. 
% and also enabling scaling up the underlying grid resolution for higher reconstruction quality
\fv{Note that, although we leverage tensor decomposition, we are not addressing a decomposition/compression problem, but a reconstruction problem based on gradient decent, since the feature grid/tensor is unknown.
In essence, our CP/VM decomposition offers low-rank regularization in the optimization, leading to high rendering quality.}
We present extensive evaluation of our approach with various settings, covering both CP and VM models, different numbers of components and grid resolutions. 
We demonstrate that all models are able to achieve realistic novel view synthesis results that are on par or better than previous state-of-the-art methods %on various datasets 
(see Fig.~\ref{fig:teaser} and Tab.~\ref{table:results}).
More importantly, our approach is of high computation and memory efficiency.
All TensoRF models can reconstruct high-quality radiance fields in 30 min; our fastest model with VM decomposition takes less than 10 min, which is significantly faster (about 100x) than NeRF and many other methods, while requiring substantially less memory than previous and concurrent voxel-based methods.
Note that, unlike concurrent works \cite{yu2021plenoxels,muller2022instant} that require unique data structures and customized CUDA kernels, our model's efficiency gains are obtained using a standard PyTorch implementation. 
%The simplicity and easy implementation of our model potentially benefits future extensions.
As far as we know, our work is the first that views radiance field modeling from a tensorial perspective and pose the problem of radiance field reconstruction as one of low-rank tensor reconstructions.

\section{Related Work}
\boldstart{Tensor decomposition.} 
% A tensor represents a multidimensional array data.
Tensor decomposition \cite{kolda2009tensor} has been studied for decades \fv{with diverse applications in vision, graphics, machine learning, and other fields \cite{panagakis2021tensor,kamal2016tensor,vasilescu2004tensortextures,deng2022constant,ballester2016tensor,ji2019survey}.}
% in various fields.
% with applications in diverse fields.
% including signal processing, computer vision, machine learning, and more.
In general, the most widely used decompositions are Tucker decomposition \cite{tucker1966some} and CP decomposition \cite{carroll1970analysis,harshman1970foundations}, both of which can be seen as generalizations of the matrix singular value decomposition (SVD). CP decomposition can also be seen as a special Tucker decomposition whose core tensor is diagonal.
By combining CP and Tucker decomposition, block term decomposition (BTD) has been proposed with its many variants \cite{de2008decompositions} and 
used in many vision and learning tasks \cite{ben2019block,ye2018learning,ye2020block}.
% \fv{used in many vision, graphics, and learning tasks \cite{kamal2016tensor,ben2019block,vasilescu2004tensortextures,deng2022constant,ballester2016tensor,ye2018learning,ye2020block,panagakis2021tensor}}. 
% used in many vision and learning tasks \fv{\cite{vasilescu2004tensortextures,kamal2016tensor,ye2018learning,ben2019block,ye2020block}}. 
In this work, we leverage tensor decomposition for radiance field modeling. 
We directly apply CP decomposition and also introduce a new vector-matrix decomposition, which can be seen as a special BTD. 

\boldstartspace{Scene representations and radiance fields.}
Various scene representations, including meshes \cite{groueix2018papier,wang2018pixel2mesh}, point clouds \cite{qi2017pointnet}, volumes \cite{ji2017surfacenet,qi2016volumetric}, implicit functions \cite{mescheder2018occupancy,peng2020convolutional}, have been extensively studied in recent years.
Many neural representations \fv{\cite{chen2018deep,zhou2018stereo,sitzmann2019deepvoxels,lombardi2019neural,bi2020deep}} are proposed for high-quality rendering or natural signal representation \cite{sitzmann2019siren,tancik2020fourfeat,liang2022coordx}.
NeRF \cite{mildenhall2020nerf} introduces radiance fields to address novel view synthesis and achieves photo-realistic quality.
This representation has been quickly extended and applied in diverse graphics and vision applications, including generative models \cite{chan2021pi,niemeyer2021giraffe}, appearance acquisition \cite{bi2020neural,boss2021nerd}, surface reconstruction \cite{wang2021neus,oechsle2021unisurf}, fast rendering\fv{ \cite{reiser2021kilonerf,yu2021plenoctrees,hedman2021baking,garbin2021fastnerf}}, appearance editing \cite{xiang2021neutex,liu2021editing}, dynamic capture \cite{li2021neural,park2021hypernerf} \fv{and generative model} \cite{Niemeyer2021CVPR,Chan2022EG3D}.  
% NeRF used pure coordinate-based MLPs to represent radiance fields. 
%  and optimize the MLPs per scene for reconstruction
While leading to realistic rendering and a compact model, NeRF with its pure MLP-based representation has known limitations in slow reconstruction and rendering. Recent methods \cite{yu2021plenoctrees,liu2020neural,hedman2021baking} have leveraged a voxel grid of features in radiance field modeling, achieving fast rendering. However, these grid-based methods still require long reconstruction time and even lead to high memory costs, sacrificing the compactness of NeRF. 
Based on feature grids, we present a novel tensorial scene representation, leveraging tensor factorization techniques, leading to fast reconstruction and compact modeling. 
% We extend the voxel-based methods to a highly compact representation based on tensor decomposition, significantly reducing the required reconstruction time and memory footprint. 
% We leverage tensor decomposition techniques to factorize a feature grid into compact vector and matrix factors for radiance field representation, significantly reducing the required reconstruction time and memory. 

Other methods design generalizable network modules trained across scenes to achieve image-dependent radiance field rendering \fv{\cite{trevithick2020grf,yu2020pixelnerf,ibrnet,Chibane2021SRF}} and fast reconstruction \cite{chen2021mvsnerf,xu2022point}. Our approach focuses on radiance field representation and only considers per-scene optimization (like NeRF). We show that our representation can already lead to highly efficient radiance field reconstruction without any across-scene generalization. We leave the extensions to generalizable settings as future work.

\boldstartspace{Concurrent work.}
The field of radiance field modeling is moving very fast and many concurrent works have appeared on arXiv as preprints over the last few months. DVGO \cite{sun2021direct} and Plenoxels \cite{yu2021plenoxels} also optimize voxel grids of (neural or SH) features for fast radiance field reconstruction. 
However, they still optimize per-voxel features directly like previous voxel-based methods, thus requiring large memory. Our approach instead factorizes the feature grid into compact components and leads to significantly higher memory efficiency.
Instant-NGP \cite{muller2022instant} uses multi-resolution hashing for efficient encoding and also leads to high compactness. 
This technique is orthogonal to our factorization-based technique; potentially, each of our vector/matrix factor can be encoded with this hashing technique and we leave such combination as future work. 
EG3D \cite{Chan2022EG3D} uses a tri-plane representation for 3D GANs; their representation is similar to our VM factorization and can be seen as a special VM version that has constant vectors. 

\section{CP and VM Decomposition}
\label{sec:vm-factor}
\begin{figure*}[t]
    \includegraphics[width=\textwidth]{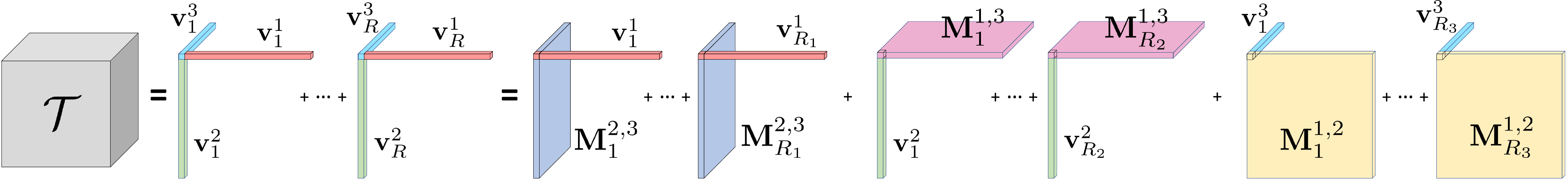}
    % \captionof{figure}{\jet{caption}}
    \caption{Tensor factorization. Left: CP decomposition (Eqn.~\ref{eqn:cpvector}), which factorizes a tensor as a sum of vector outer products.  Right: our vector-matrix decomposition (Eqn.~\ref{eqn:vmrrr}), which factorizes a tensor as a sum of vector-matrix outer products.}
    \label{fig:tensor-decomp}\vspace{-5mm}
\end{figure*}

We factorize radiance fields into compact components for scene modeling.
To do so, we apply both the classic CP decomposition and a new vector-matrix (VM) decomposition; both are illustrated in Fig.~\ref{fig:tensor-decomp}.
% While standard tensor decompositions can be applied, we introduce a new vector-matrix decomposition, facilitating the radiance field reconstruction and rendering.
We now discuss both decompsitions with an example of a 3D (3rd-order) tensor. We will introduce how to apply tensor factorization in radiance field modeling (with a 4D tensor) in Sec.~\ref{sec:tensorf}.
% We now review the classical CP decomposition and present our decomposition for the case of a 3D (3rd-order) tensor; both are illustrated in Fig.~\ref{fig:tensor-decomp}. We will introduce how to apply the decomposition in radiance field modeling (with a 4D tensor) in Sec.~\ref{sec:tensorf}.

\boldstartspace{CP decomposition.} 
Given a 3D tensor $\Tensor \in \RRR^{I\times J\times K}$, CP decomposition factorizes it into a sum of outer products of vectors (shown in Fig.~\ref{fig:tensor-decomp}):
% Given a 3D tensor $\Tensor \in \RRR^{I\times J\times K}$, CP decomposition factorizes this tensor into a sum of outer products of vectors (shown in Fig.~\ref{fig:tensor-decomp} left):
\begin{equation}
    \Tensor = \sum_{r=1}^R \Vector_r^{1} \OuterP \Vector_r^{2} \OuterP \Vector_r^{3}
    \label{eqn:cpvector}
\end{equation}
where $\Vector_r^{1} \OuterP \Vector_r^{2} \OuterP \Vector_r^{3}$ corresponds to a rank-one tensor component, and $\Vector_r^{1} \in \RRR^{I}$, $\Vector_r^{2} \in \RRR^{J}$, and $\Vector_r^{3} \in \RRR^{K}$ are factorized vectors of the three modes for the $r$th component. Superscripts denote the modes of each factor; $\OuterP$ represents the outer product. Hence, each tensor element $\Tensor_{ijk}$ is a sum of scalar products:
\begin{equation}
    \Tensor_{ijk} = \sum_{r=1}^R \Vector_{r,i}^{1} \Vector_{r,j}^{2} \Vector_{r,k}^{3}
    \label{eqn:cpelement}
\end{equation}
where $i$, $j$, $k$ denote the indices of the three modes. 
% and $\Vector_{r,i}^{1}$, $\Vector_{r,j}^{2}$,$\Vector_{r,k}^{3}$ are indexed scalar values from the mode vectors.

CP decomposition factorizes a tensor into multiple vectors, expressing multiple compact rank-one components. 
CP can be directly applied in our tensorial radiance field modeling and generate high-quality results (see Tab.~\ref{table:results}). However, because of too high compactness, CP decomposition can require many components to model complex scenes, leading to high computational costs in radiance field reconstruction.
Inspired by block term decomposition (BTD), we present a new VM decomposition, leading to more efficient radiance field reconstruction.

\boldstartspace{Vector-Matrix (VM) decomposition.} Unlike CP decomposition that utilizes pure vector factors, VM decomposition factorizes a tensor into multiple vectors and matrices as shown in Fig.~\ref{fig:tensor-decomp} right. This is expressed by
\begin{equation}
    \Tensor = \sum_{r=1}^{R_1} \Vector_r^{1} \OuterP \Matrix_r^{2,3} + \sum_{r=1}^{R_2}\Vector_r^{2} \OuterP \Matrix_r^{1,3} + \sum_{r=1}^{R_3}\Vector_r^{3} \OuterP \Matrix_r^{1,2} 
    \label{eqn:vmrrr}
\end{equation}
where $\Matrix_r^{2,3} \in \RRR^{J\times K}$, $\Matrix_r^{1,3} \in \RRR^{I\times K}$, $\Matrix_r^{1,2} \in \RRR^{I\times J}$ are matrix factors for two (denoted by superscripts) of the three modes.
For each component, we relax its two mode ranks to be arbitrarily large, while restricting the third mode to be rank-one; e.g., for component tensor $\Vector_r^{1} \OuterP \Matrix_r^{2,3}$, its mode-1 rank is 1, and its mode-2 and mode-3 ranks can be arbitrary, depending on the rank of the matrix $\Matrix_r^{2,3}$.
In general, instead of using separate vectors in CP, we combine every two modes and represent them by matrices, allowing each mode to be adequately parametrized with a smaller number of components. $R_1$, $R_2$, $R_3$ can be set differently and should be chosen depending on the complexity of each mode.
% In fact, the classical rank-(L,L,1) BTD can be seen as a reduced version of our decomposition, where $R_1=R_2=0$ (only two fixed modes are chosen for the matrix factors) and the rank of matrix factors is reduced and fixed to L.
Our VM decomposition can be seen as a special case of general BTD. 

Note that, each of our component tensors has more parameters than a component in CP decomposition. 
While this leads to lower compactness, a VM component tensor can express more complex high-dimensional data than a CP component, thus reducing the required number of components when modeling the same complex function. 
% While this leads to lower compactness, our decomposition requires significantly fewer components by replacing the two of vectors mode with learnable matrix, leading to efficient computation in ray marching.
On the other hand, VM decomposition is still of very high compactness, reducing memory complexity from $\mathcal{O}(N^3)$ to $\mathcal{O}(N^2)$, compared to dense grid representations.

% Our decomposition can potentially be extended to general tensors of arbitrary order (no less than 3) and used in other applications.
\boldstartspace{Tensor for scene modeling.} 
In this work, we focus on the task of modeling and reconstructing radiance fields.
% In this work, we focus on the task of modeling and reconstructing radiance fields (which includes a 4D tensor; see details in Sec.~\ref{sec:tensorf}).
% While our decomposition can potentially be extended to general tensors of arbitrary order (no less than 3). We focus on this version of 3D tensors and a similar version for 4D tensors (see details in Sec.~\ref{sec:tensorf}) for radiance field modeling in this paper.
In this case, the three tensor modes correspond to XYZ axes, and we thus directly denote the modes with XYZ to make it intuitive.
Meanwhile, in the context of 3D scene representation, we consider $R_1=R_2=R_3=R$ for most of scenes, reflecting the fact that a scene can distribute and appear equally complex along its three axes.
Therefore, Eqn.~\ref{eqn:vmrrr} can be re-written as
\begin{equation}
    \Tensor = \sum_{r=1}^{R} \Vector_r^{X} \OuterP \Matrix_r^{Y,Z} + \Vector_r^{Y} \OuterP \Matrix_r^{X,Z} + \Vector_r^{Z} \OuterP \Matrix_r^{X,Y}
    \label{eqn:vmr}
\end{equation}
In addition, to simplify notation and the following discussion in later sections, we also denote the three types of component tensors as $\Comp^X_r=\Vector_r^X \OuterP \Matrix_r^{YZ}$, $\Comp^Y_r=\Vector_r^Y \OuterP \Matrix_r^{XZ}$, and $\Comp^Z_r=\Vector_r^Z \OuterP \Matrix_r^{XY}$; here the superscripts XYZ of $\Comp$ indicate different types of components.
With this, a tensor element $\Tensor_{ijk}$ is expressed as
\begin{align}
    \Tensor_{ijk} = & \sum_{r=1}^{R} \sum_{m} \Comp_{r,ijk}^m
    \label{eqn:vmrelement}
\end{align}
% \begin{align}
%     \Tensor_{ijk} = & \sum_{r=1}^{R} \Vector_{r,i}^{X} \Matrix_{r,jk}^{YZ} + \Vector_{r,j}^{Y} \Matrix_{r,ik}^{XZ}+ \Vector_{r,k}^{Z} \Matrix_{r,ij}^{XY} \notag \\
%                   = & \sum_{r=1}^{R} \Comp^X_{r,ijk} + \Comp^Y_{r,ijk}+ \Comp^Z_{r,ijk} \notag \\
%                   = & \sum_{r=1}^{R} \sum_{m\in XYZ} \Comp_{r,ijk}^m
%     \label{eqn:vmrelement}
% \end{align}
where $m\in XYZ$, $\Comp^X_{r,ijk}=\Vector_{r,i}^X \Matrix_{r,jk}^{YZ}$, $\Comp^Y_{r,ijk}=\Vector_{r,j}^Y \Matrix_{r,ik}^{XZ}$, and $\Comp^Z_{r,ijk}=\Vector_{r,k}^Z \Matrix_{r,ij}^{XY}$. Similarly, we can also denote a CP component as $\Comp^{\gamma}=\Vector_r^X \OuterP \Vector_r^Y \OuterP \Vector_r^Z$, and Eqn.~\ref{eqn:vmrelement} can also express CP decomposition by considering $m=\gamma$, where the summation over $m$ can be removed.
 
% Unlike rank-(L,L,1) BTD, we do not constrain the rank of a matrix factor and we alternate the tensor modes for the matrix in different components, facilitating the task of radiance field reconstruction. 

% We alternate the two modes for the matrix factor to allow each mode to be adequately parametrized. In most cases, we also set $R_1=R_2=R_3=R$; in the context of 3D scene representation, this reflects the fact that a scene can distribute and appear equally complex along its three axes. This leads to a reduced version of Eqn.~\ref{eqn:vmrrr}
% \begin{equation}
%     \Tensor = \sum_{r=1}^{R} \Vector_r^{1} \OuterP \Matrix_r^{2,3} + \Vector_r^{2} \OuterP \Matrix_r^{1,3} + \Vector_r^{3} \OuterP \Matrix_r^{1,2}, 
%     \label{eqn:vmr}
% \end{equation}
% and its element-wise format is expressed by
% \begin{equation}
%     \Tensor_{ijk} = \sum_{r=1}^{R} \Vector_{r,i}^{1} \ScalarP \Matrix_{r,jk}^{2,3} + \Vector_{r,j}^{2} \ScalarP \Matrix_{r,ik}^{1,3} + \Vector_{r,k}^{3} \ScalarP \Matrix_{r,ij}^{1,2}, 
%     \label{eqn:vmrelement}
% \end{equation}

\section{Tensorial Radiance Field Representation}
\label{sec:tensorf}
We now present our Tensorial Radiance Field Representation (TensoRF). For simplicity, we focus on presenting TensoRF with our VM decomposition. CP decomposition is simpler and its decomposition equations can be directly applied with minimal modification (like Eqn.~\ref{eqn:vmrelement}).

\subsection{Feature grids and radiance field.}
\label{sec:grids}
Our goal is to model a radiance field, which is essentially a function that maps any 3D location $\Pos$ and viewing direction $\Dir$ to its volume density $\Dens$ and view-dependent color $\Rad$, supporting differentiable ray marching for volume rendering.
We leverage a regular 3D grid $\Grid$ with per-voxel multi-channel features to model such a function. %(shown in Fig.~\ref{fig:grids})
We split it (by feature channels) into a geometry grid $\Grid_\Dens$ and an appearance grid $\Grid_\Rad$, separately modelling the volume density $\Dens$ and view-dependent color $\Rad$.

Our approach supports various types of appearance features in $\Grid_\Rad$, depending on a pre-selected function $\ShadFunc$ that coverts an appearance feature vector and a viewing direction $\Dir$ to color $\Rad$. For example, $\ShadFunc$ can be a small MLP or spherical harmonics (SH) functions, where $\Grid_\Rad$ contains neural features and SH coefficients respectively.
We show that both MLP and SH functions work well with our model (see Tab.\ref{table:results}).
On the other hand, we consider a single-channel grid $\Grid_\Dens$, whose values represent volume density directly, without requiring an extra converting function. The continuous grid-based radiance field can be written as
\begin{equation}
    \Dens, \Rad = \Grid_\Dens(\Pos), \ShadFunc(\Grid_\Rad(\Pos),\Dir)
    \label{eqn:rf}
\end{equation}
where $\Grid_\Dens(\Pos)$, $\Grid_\Rad(\Pos)$ represent the trilinearly interpolated features from the two grids at location $\Pos$. 
We model $\Grid_\Dens$ and $\Grid_\Rad$ as factorized tensors.
% , and factorize them for modeling.

\subsection{Factorizing radiance fields}
While $\Grid_\Dens \in \RRR^{\DimIJK}$ is a 3D tensor, $\Grid_\Rad \in \RRR^{\DimIJK\times \DimCh}$ is a 4D tensor. Here $I$, $J$, $K$ correspond to the resolutions of the feature grid along the X, Y, Z axes, and $\DimCh$ is the number of appearance feature channels. 

We factorize these radiance field tensors to compact components. In particular, with the VM decomposition. %, as illustrated in Fig.~\ref{fig:grids}.
% In this section, we use additional subscripts $\Dens$, $\Rad$ if necessary to denote the variables for the two grids.
% In particular, the 3D geometry tensor $\Grid_\Dens$ is factorized into by vectors $\Vector_{\Dens, r}^{X}$, $\Vector_{\Dens, r}^{Y}$, $\Vector_{\Dens, r}^{Z}$, and matrices $\Matrix_{\Dens,r}^{YZ}$, $\Matrix_{\Dens,r}^{XZ}$, $\Matrix_{\Dens,r}^{XY}$, for $r=1,..,R_\Dens$. Here, since the modes of our tensors correspond to the coordinate frame, we directly use $X$, $Y$, $Z$ to denote the three modes. These vector and matrix factors essentially describe the spatial variations of the scene geometry (volume density) along their corresponding axes. 
the 3D geometry tensor $\Grid_\Dens$ is factorized as 
% Similar to Eqn.~\ref{eqn:vmrelement}, each grid element is computed by
% \begin{equation}
%     \Grid_{\Dens,ijk} = \sum_{r=1}^{R} \Vector_{\Dens r,i}^{X} \Matrix_{\Dens r,jk}^{Y,Z} + \Vector_{\Dens r,j}^{Y} \Matrix_{\Dens r,ik}^{X,Z} + \Vector_{\Dens r,k}^{Z} \Matrix_{\Dens r,ij}^{X,Y}. 
%     \label{eqn:dense-vmr}
% \end{equation}
\begin{align}
    \Grid_{\Dens} = \sum_{r=1}^{R_\Dens} \Vector_{\Dens, r}^{X} \OuterP \Matrix_{\Dens,r}^{YZ} + \Vector_{\Dens,r}^{Y} \OuterP \Matrix_{\Dens,r}^{XZ} + \Vector_{\Dens,r}^{Z} \OuterP \Matrix_{\Dens,r}^{XY}
                 = \sum_{r=1}^{R_\Dens} \sum_{m\in XYZ} \Comp^m_{\Dens, r}
    \label{eqn:dense-vmr}
\end{align}

% We also denote tensor components $\Comp_{\Dens,r}^X=\Vector_{\Dens, r}^{X} \OuterP \Matrix_{\Dens,r}^{YZ}$, $\Comp_{\Dens,r}^Y=\Vector_{\Dens, r}^{Y} \OuterP \Matrix_{\Dens,r}^{XZ}$, $\Comp_{\Dens,r}^Z=\Vector_{\Dens, r}^{Z} \OuterP \Matrix_{\Dens,r}^{XY}$.

% The appearance tensor $\Grid_\Rad$ has an additional mode corresponding to the feature channel dimension. Note that, compared to XYZ modes, this mode is often of a lower rank, since its dimension is generally lower. Therefore, we do not combine this mode with other modes in matrix factors and instead only use vectors, denoted by $\AppVec_r$, for this mode in the factorization. 
The appearance tensor $\Grid_\Rad$ has an additional mode corresponding to the feature channel dimension. Note that, compared to the XYZ modes, this mode is often of lower dimension, leading to a lower rank. Therefore, we do not combine this mode with other modes in matrix factors and instead only use vectors, denoted by $\AppVec_r$, for this mode in the factorization. 
% Specifically, $\Grid_\Rad$ is factorized into $\Vector_{\Rad, r}^{X}$, $\Vector_{\Rad, r}^{Y}$, $\Vector_{\Rad, r}^{Z}$, $\Matrix_{\Rad,r}^{YZ}$, $\Matrix_{\Rad,r}^{XZ}$, $\Matrix_{\Rad,r}^{XY}$, for $r=1,..,R_\Rad$, and $\AppVec_r$, for $r=1,..,3R_\Rad$. Note that, we need to have $3R_\Rad$ vectors $\AppVec_r$ to match the total number of components. In particular, $\Grid_\Rad$ is expressed as
Specifically, $\Grid_\Rad$ is factorized as
% \begin{equation}
%     \Grid_{\Rad,ijk} = \sum_{r=1}^{R} \Vector_{\Rad r,i}^{X} \Matrix_{\Rad r,jk}^{YZ} \AppVec_{r,p} + \Vector_{\Rad r,j}^{Y} \Matrix_{\Rad r,ik}^{XZ} \AppVec_{r,p} + \Vector_{\Rad r,k}^{Z} \Matrix_{\Rad r,ij}^{XY} \AppVec_{r,p}. 
%     \label{eqn:rad-vmr}
% \end{equation}
% \begin{align}
%     \Grid_{\Rad} = \sum_{r=1}^{R_\Rad}& \Vector_{\Rad, r}^{X} \OuterP \Matrix_{\Rad,r}^{YZ} \OuterP \AppVec_{3r-2} + \Vector_{\Rad,r}^{Y} \OuterP \Matrix_{\Rad,r}^{XZ} \OuterP \AppVec_{3r-1} \notag \\
%     &+ \Vector_{\Rad,r}^{Z} \OuterP \Matrix_{\Rad,r}^{XY} \OuterP \AppVec_{3r}. 
%     \label{eqn:rad-vmr}
% \end{align}
\begin{align}
    \Grid_{\Rad} = \sum_{r=1}^{R_\Rad}& \Vector_{\Rad, r}^{X} \OuterP \Matrix_{\Rad,r}^{YZ} \OuterP \AppVec_{3r-2} + \Vector_{\Rad,r}^{Y} \OuterP \Matrix_{\Rad,r}^{XZ} \OuterP \AppVec_{3r-1}
    + \Vector_{\Rad,r}^{Z} \OuterP \Matrix_{\Rad,r}^{XY} \OuterP \AppVec_{3r} \notag \\
    =\sum_{r=1}^{R_\Rad} &\Comp_{\Rad,r}^X \OuterP \AppVec_{3r-2} +\Comp_{\Rad,r}^Y \OuterP \AppVec_{3r-1} +\Comp_{\Rad,r}^Z \OuterP \AppVec_{3r}
    \label{eqn:rad-vmr}
\end{align}
Note that, we have $3R_\Rad$ vectors $\AppVec_r$ to match the total number of components.

% Figure~\ref{fig:grids} illustrates all the factors that we use to model tensorial radiance fields.
Overall, we factorize the entire tensorial radiance field into $3R_\Dens+3R_\Rad$ matrices ($\Matrix_{\Dens,r}^{YZ}$,$...$,$\Matrix_{\Rad,r}^{YZ}$,...) and $3R_\Dens+6R_\Rad$ vectors ($\Vector_{\Dens, r}^{X}$,...,$\Vector_{\Rad, r}^{X}$,...,$\AppVec_r$). 
% Note that, the total number of parameters of our model can be computed by $3I^2(R_\Rad+R_\sigma)+3I(R_\Rad+R_\sigma)+3R_\Rad P$
In general, we adopt $R_\Dens \ll I, J, K$ and $R_\Rad \ll I, J, K$, leading to a highly compact representation that can encode a high-resolution dense grid.
In essence, the XYZ-mode vector and matrix factors, $\Vector_{\Dens, r}^{X}$, $\Matrix_{\Dens,r}^{YZ}$, $\Vector_{\Rad, r}^{X}$, $\Matrix_{\Rad,r}^{YZ}$, ..., describe the spatial distributions of the scene geometry and appearance along their corresponding axes. 
On the other hand, the appearance feature-mode vectors $\AppVec_r$ express the global appearance correlations.
By stacking all $\AppVec_r$ as columns together, we have a $P\times 3R_\Rad$ matrix $\AppMat$; this matrix $\AppMat$ can also be seen as a global appearance dictionary that abstracts the appearance commonalities across the entire scene. 

% These vector and matrix factors essentially describe the spatial variations of the scene geometry (volume density) along their corresponding axes. 
% Each tensor element $\Grid_{\Dens,ijk}$, $\Grid_{\Dens,ijkp}$ are computed by
% \begin{align}
%     \Grid_{\Dens,ijk} =  \sum_{r=1}^{R_\Dens} & \Vector_{\Dens,r,i}^{X} \Matrix_{\Dens,r,jk}^{YZ} + \Vector_{\Dens,r,j}^{Y} \Matrix_{\Dens,r,ik}^{XZ} \notag \\
%     & + \Vector_{\Dens,r,k}^{Z} \Matrix_{\Dens,r,ij}^{XY} \label{eqn:dens-vmrelement} \\
%     \Grid_{\Rad,ijkp} =  \sum_{r=1}^{R_\Rad} & \Vector_{\Rad,r,i}^{X} \Matrix_{\Rad,r,jk}^{YZ} \AppVec_{3r-2,p} + \Vector_{\Rad,r,j}^{Y} \Matrix_{\Rad,r,ik}^{XZ} \AppVec_{3r-1,p} \notag\\
%     & + \Vector_{\Rad r,k}^{Z} \Matrix_{\Rad r,ij}^{XY} \AppVec_{3r,p}. 
%     \label{eqn:rad-vmrelement}
% \end{align}
% Here $i$,$j$,$k$ are indices for the XYZ modes and $p$ is the feature channel index.

\begin{figure*}[t]
    \includegraphics[width=\textwidth]{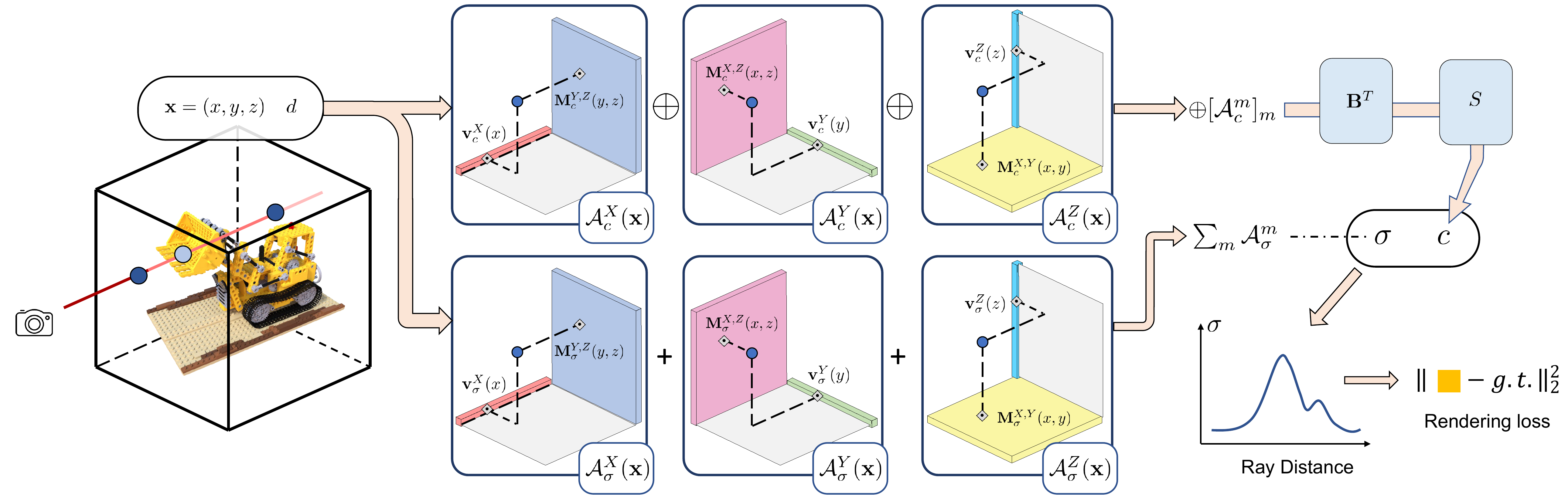}
    % \captionof{figure}{\jet{caption}}
    \caption{TensoRF (VM) reconstruction and rendering. We model radiance fields as tensors using a set of vectors ($\Vector$) and matrices ($\Matrix$), which describe the scene along their corresponding (XYZ) axes and are used for computing volume density $\Dens$ and view-dependent color $\Rad$ in differentiable ray marching.
    % These factors are used for computing volume density $\Dens$ and view-dependent color $\Rad$ separately, allowing rendering images with differentiable ray marching.   
    For each shading location $\Pos=(x,y,z)$, we use linearly/bilinearly sampled values from the vector/matrix factors to efficiently compute the corresponding trilinearly interpolated values ($\Comp(\Pos)$) of the tensor components. The density component values ( $\Comp_\Dens(\Pos)$) are summed 
    to get the volume density ($\Dens$) directly. 
    % to compute the volume density $\Dens$. 
    The appearance values ($\Comp_\Rad(\Pos)$) are concatenated into a vector ($\oplus[\Comp_\Rad^m(x)]_m$) that is then multiplied by an appearance matrix $\AppMat$ and sent to the decoding function $S$ for RGB color ($\Rad$) regression. 
    % (an MLP or spherical harmonics function) 
    % leading to an appearance vector that a shading function $S$ (an MLP or spherical harmonics function) converts to the view-dependent color $\Rad$. We optimize our tensorial radiance fields using a standard L2 rendering loss. 
    }
    \label{fig:pipeline}\vspace{-5mm}
\end{figure*}

\subsection{Efficient feature evaluation.}
% Radiance field rendering is based on ray marching that requires sampling a lot of 3D locations and computing their density values and view-dependent colors.
% This desires our representation to efficiently compute the volume features at any 3D locations.
% Note that, with our factorized representation, any element value of the original tensors can already be evaluated at very low costs; as shown in Eqn.~\ref{eqn:vmrelement},  computing one tensor element only requires indexing one value per (vector or matrix) factor and combining them with simple multiplications and additions.
% As is modeled by the sum of outer products, our model naturally supports low-cost direct evaluation of recovering each tensor element. 

% We now introduce how to efficiently decode features with our factorized representation.
Our factorization-based model can compute each voxel's feature vector at low costs, only requiring one value per XYZ-mode vector/matrix factor.
We also enable efficient trilinear interpolation for our model, leading to a continuous field.  %enhancing the spatial coherence and 

\boldstartspace{Direct evaluation.}
With VM factorization, a density value $\Grid_{\Dens,ijk}$ of a single voxel at indices $ijk$ can be directly and efficiently evaluated by following Eqn.~\ref{eqn:vmrelement}:
\begin{align}
    \Grid_{\Dens,ijk} =  \sum_{r=1}^{R_\Dens} \sum_{m\in XYZ} \Comp^m_{\Dens,r,ijk}
    \label{eqn:dens-vmrelement}
\end{align}
Here, computing each $\Comp^m_{\Dens,r,ijk}$ only requires indexing and multiplying two values from its corresponding vector and matrix factors.

As for the appearance grid $\Grid_{\Rad}$, we always need to compute a full 
$\DimCh$-channel feature vector, which the shading function $\ShadFunc$ requires as input, corresponding to a 1D slice of $\Grid_{\Rad}$ at fixed XYZ indices $ijk$:
% This feature vector, simply denoted by $\Grid_{\Rad,ijk}$, can be computed in a similar way by
\begin{align}
    \Grid_{\Rad,ijk} =  \sum_{r=1}^{R_\Rad} \Comp^X_{\Rad,r,ijk} \AppVec_{3r-2} + \Comp^Y_{\Rad,r,ijk} \AppVec_{3r-1}
     + \Comp^Z_{\Rad,r,ijk} \AppVec_{3r}
    \label{eqn:rad-vmrslice}
\end{align}
Here, there's no additional indexing for the feature mode, since we compute a full vector.
We further simplify Eqn.~\ref{eqn:rad-vmrslice} by re-ordering the computation.
For this, we denote $\oplus [\Comp^m_{\Rad,ijk}]_{m,r}$ as the vector that stacks all $\Comp^m_{\Rad,r,ijk}$ values for $m=X,Y,Z$ and $r=1,...,R_\Rad$, which is a vector of $3R_\Rad$ dimensions; $\oplus$ can also be considered as the concatenation operator that concatenates all scalar values (1-channel vectors) into a $3R_\Rad$-channel vector in practice.
Using matrix $\AppMat$ (introduced in Sec.~\ref{sec:grids}) that stacks all $\AppVec_r$, Eqn.~\ref{eqn:rad-vmrslice} is equivalent to a matrix vector product:
\begin{align}
    \Grid_{\Rad,ijk} =  \AppMat (\oplus[\Comp^m_{\Rad,ijk}]_{m,r})
    \label{eqn:rad-vmrfeature}
\end{align}
Note that, Eqn.~\ref{eqn:rad-vmrfeature} is not only formally simpler but also leads to a simpler implementation in practice.
Specifically, when computing a large number of voxels in parallel, we first compute and concatenate $\Comp^m_{\Rad,r,ijk}$ for all voxels as column vectors in a matrix and then multiply the shared matrix $\AppMat$ once. 

% Note that, Eqn.~\ref{eqn:rad-vmrfeature} is not only simpler in its format but also leads to more efficient computation than Eqn.~\ref{eqn:rad-vmrslice} in practice.
% Specifically, when computing a large number of feature vectors in parallel, following Eqn.~\ref{eqn:rad-vmrfeature} allows for computing and concatenating $\Comp^m_{\Rad,r,ijk}$ first and then multiplying the shared matrix $\AppMat$ once. 

% Note that, as shown in Eqn.~\ref{eqn:dens-vmrelement} and \ref{eqn:rad-vmrelement}, our model already allows for each tensor element to be evaluated at very low costs, requiring indexing only one value per (vector or matrix) factor and combining them with simple multiplications and additions.
% % The original tensors represent discrete feature data. 
% We further enable efficient trilinear interpolation for our model, enhancing the spatial coherence and leading to a continuous field. 
% Moreover, since the shading function ($\ShadFunc$, discussed in Sec.~\ref{sec:grids}) always requires a full $\DimCh$-channel feature vector to compute view-dependent color, 
% we also introduce a simple technique to efficiently obtain a full feature vector, without requiring computing individual channels separately as Eqn.~\ref{eqn:rad-vmrelement}.

% To ease the following discussion, we also denote tensor components $\Comp^X_r=\Vector_r^X \OuterP \Matrix_r^{YZ}$, $\Comp^Y_r=\Vector_r^Y \OuterP \Matrix_r^{XZ}$, and $\Comp^Z_r=\Vector_r^Z \OuterP \Matrix_r^{XY}$; here the superscripts XYZ over $\Comp$ are for separating different components.

%Instead of using tensor values directly, 
\boldstartspace{Trilinear interpolation.}
We apply trilinear interpolation to model a continuous field.
Na\"{i}vely achieving trilinear interpolation is costly, as it requires evaluation of 8 tensor values and interpolating them, increasing computation by a factor of 8 compared to computing a single tensor element.
However, we find that trilinearly interpolating a component tensor is naturally equivalent to interpolating its vector/matrix factors linearly/bilinearly for the corresponding modes, thanks to the beauty of linearity of the trilinear interpolation and the outer product.
% However, naively achieving trilinear interpolation is costly, which requires recovering 8 corner elements separately and interpolating them, increasing 8x computation and memory costs compared to computing a single tensor element.
% However, we find that trilinearly interpolating a component tensor is naturally equivalent to interpolating its vector/matrix factors linearly/bilinearly for the corresponding modes, thanks to the beauty of linearity of the trilinear interpolation and our outer product based factorization.
% Naively achieving trilinear interpolation for our tensorial model requires first recovering 8 corner tensor elements (using Eqn.~\ref{eqn:dens-vmrelement} and \ref{eqn:rad-vmrelement}) and then interpolating them trilinearly. This increases 8x computation and memory costs compared to computing a single tensor element. 
% However, we find that trilinearly interpolating a tensor data along its three modes is naturally equivalent to interpolating its vector/matrix factors linearly/bilinearly for the corresponding modes.

For example, given a component tensor $\Comp^X_r=\Vector_r^X \OuterP \Matrix_r^{YZ}$ with its each tensor element $\Comp_{r,ijk}=\Vector_{r,i}^X \Matrix_{r,jk}^{YZ}$, we can compute its interpolated values as:
\begin{equation}
    \Comp^X_r(\Pos) = \Vector_r^X(x)\Matrix_{r}^{YZ}(y,z)
    \label{eqn:xinter}
\end{equation}
where $\Comp^X_r(\Pos)$ is $\Comp_r$'s trilinearly interpolated value at location $\Pos=(x,y,z)$ in the 3D space, $\Vector_r^X(x)$ is $\Vector_r^X$'s linearly interpolated value at $x$ along X axis, and $\Matrix_{r}^{YZ}(y,z)$ is $\Matrix_{r}^{YZ}$'s bilinearly interpolated value at $(y,z)$ in the YZ plane. Similarly, we have $\Comp^Y_r(\Pos) = \Vector_r^Y(y)\Matrix_{r}^{XZ}(x,z)$ and $\Comp^Z_r(\Pos) = \Vector_r^Z(z)\Matrix_{r}^{XY}(x,y)$ (for CP decomposition $\Comp^\gamma_r(\Pos) = \Vector_r^X(x)\Vector_r^Y(y)\Vector_r^Z(z)$ is also valid).
% \begin{align}
%     \Comp^Y_r(\Pos) &= \Vector_r^Y(y)\Matrix_{r}^{XZ}(x,z),\label{eqn:yinter} \\
%     \Comp^Z_r(\Pos) &= \Vector_r^Z(z)\Matrix_{r}^{XY}(x,y).\label{eqn:zinter}
% \end{align}
Thus, trilinearly interpolating the two grids is expressed as
\begin{align}
    \Grid_{\Dens}(\Pos) &= \sum_r \sum_m \Comp^m_{\Dens,r}(\Pos) 
    \label{eqn:dens-vmrinter}
    \\
    \Grid_{\Rad}(\Pos) &=  \AppMat(\oplus[\Comp^m_{\Rad,r}(\Pos)]_{m,r})
    \label{eqn:rad-vmrinter}
\end{align}
% Therefore, we can efficiently compute trilinearly interpolation at a low cost similar to computing a single tensor element, using a few linear and bilinear interpolations on the vector and matrix factors. 
These equations are very similar to Eqn.~\ref{eqn:dens-vmrelement} and \ref{eqn:rad-vmrfeature}, simply replacing the tensor elements with interpolated values.
We avoid recovering 8 individual tensor elements for trilinear interpolation and instead directly recover the interpolated value, leading to low computation and memory costs at run time.
% This reflects the low cost of our trilinear interpolation, which is similar to the trilinear interpolation on a standard voxel grid. 
%Therefore, our tensorial representation models a continuous radiance field.
% with which we can efficiently compute continuous volume density $\Grid_{\Dens}(\Pos)$ and appearance feature $\Grid_{\Rad}(\Pos)$ at any 3D location $\Pos$ inside the grid.
% Therefore, our tensorial representation models a continuous radiance field, with which we can efficiently compute continuous volume density $\Grid_{\Dens}(\Pos)$ and appearance feature $\Grid_{\Rad}(\Pos)$ at any 3D location $\Pos$ inside the grid.
% For example, given a component tensor $\Comp_r=\Vector_r^X \OuterP \Matrix_r^{YZ}$, we already know $\Comp_{r,ijk}=\Vector_{r,i}^X \OuterP \Matrix_{r,jk}^{YZ}$ by definition, and now we also have
% \begin{equation}
%     \Comp_r(\Pos) = \Vector_r^X(x)\Matrix_{r}^{YZ}(y,z),
% \end{equation}
% where $\Comp_r(\Pos)$ is $\Comp_r$'s trilinearly interpolated value at location $\Pos=(x,y,z)$ in 3D space, $\Vector_r^X(x)$ is $\Vector_r^X$'s linear interpolated value at $x$ along the X axis, and $\Matrix_{r}^{YZ}(y,z)$ is $\Matrix_{r}^{YZ}$'s bilinearly interpolated value at $(y,z)$ in the YZ plane. Therefore, we can efficiently compute trilinearly interpolation with a few linear and bilinear interpolation on the vector and matrix factors.

\subsection{Rendering and reconstruction.}
\label{sec:rendering_reconstruction}
Equations~\ref{eqn:rf}, \ref{eqn:xinter}--\ref{eqn:rad-vmrinter} describe the core components of our model. By combining Eqn.~\ref{eqn:rf},\ref{eqn:dens-vmrinter},\ref{eqn:rad-vmrinter}, our factorized tensorial radiance field can be expressed as
\begin{equation}
    \Dens, \Rad=\sum_{r}\sum_m \Comp^m_{\Dens,r}(\Pos)\:,\:
    \ShadFunc(\AppMat(\oplus[\Comp^m_{\Rad,r}(\Pos)]_{m,r}), \Dir)
\end{equation}
i.e., we obtain continuous volume density and view-dependent color given any 3D location and viewing direction. 
This allows for high-quality radiance field reconstruction and rendering.
Note that, this equation is general and describes TensoRF with both CP and VM decomposition.
Our full pipeline of radiance field reconstruction and rendering with VM decomposition is illustrated in Fig.~\ref{fig:pipeline}.

\boldstartspace{Volume rendering.} 
To render images, we use differentiable volume rendering, following NeRF \cite{mildenhall2020nerf}. Specifically, for each pixel, we march along a ray, sampling $Q$ shading points along the ray and computing the pixel color by
% This allows for differentiable volume rendering as done in NeRF \cite{mildenhall2020nerf}, achieved by ray marching, which marches a ray through a pixel center, samples $N$ shading points, and computes the pixel color $\Color$ by  
\begin{equation}
        \Color =  \sum_{q=1}^Q \Trans_q (1-\exp (-\Dens_q \Step_q)) \Rad_q, \ 
        \Trans_q = \exp (-\sum_{p=1}^{q-1} \Dens_p \Step_p )
    \label{eq:raymarching}
\end{equation}
Here, $\Dens_q$, $\Rad_q$ are the corresponding density and color computed by our model at their sampled locations $\Pos_q$; $\Delta_q$ is the ray step size and $\Trans_q$ represents transmittance. 

\boldstartspace{Reconstruction.}
Given a set of multi-view input images with known camera poses, our tensorial radiance field is optimized per scene via gradient descent, minimizing an L2 rendering loss, using only the ground truth pixel colors as supervision.
% We use only a L2 rendering loss for the reconstruction:
% \begin{equation}
%     \mathcal{L}=\|\Color-\hat{\Color}\|_2^2,
% \end{equation}
% where $\hat{\Color}$ is the ground truth color from the captured image.
% Additionally, we also apply an L1 sparsity loss.
Our radiance field is explained by tensor factorization and modeled by a set of global vectors and matrices as basis factors that correlate and regularize the entire field in the optimization.
However, this can sometimes lead to overfitting and local minima issues in gradient descent,
leading to outliers or noises in regions with fewer observations.
% where the factors mainly explain regions that are densely covered by rays and generate outliers or noise in regions with fewer observations.
% Inspired by classical compressive sensing techniques, we utilize standard regularization terms, including an L1 norm sparsity loss and a TV (total variation) loss on our vector and matrix factors, which effectively address these issues.
We utilize standard regularization terms that are commonly used in compressive sensing, including an L1 norm loss and a TV (total variation) loss on our vector and matrix factors, which effectively address these issues.
We find that only applying the L1 sparsity loss is adequate for most datasets. However, for real datasets that have very few input images (like LLFF\cite{llff}) or imperfect capture conditions (like Tanks and Temples \cite{Knapitsch2017,liu2020neural} that has varying exposure and inconsistent masks), a TV loss is more efficient than the L1 norm loss.

% Note that our radiance field is explained by tensor factorization and modeled by a set of global vectors and matrices as basis factors that correlate and regularize the entire field. Empirically, we find that this leads to robust solutions even without additional smoothness or regularization loss terms used in previous or concurrent voxel-based works.

% Because our model is compact and efficient to evaluate, our optimization converges very fast usually in less than 30k steps. 
To further improve quality and avoid local minima, we apply coarse-to-fine reconstruction. Unlike previous coarse-to-fine techniques that require unique subdivisions on their sparse chosen sets of voxels, our coarse-to-fine reconstruction is simply achieved by linearly and bilinearly upsampling our XYZ-mode vector and matrix factors.

\section{Implementation details}

\label{sec:impl}
We briefly discuss our implementation; please refer to the appendix for more details.
We implement our TensoRF using PyTorch \cite{pytorch}, without customized CUDA kernels.
We implement the feature decoding function $\ShadFunc$ as either an MLP or SH function and use $P=27$ features for both. For SH, this corresponds to 3rd-order SH coefficients with RGB channels. For neural features, we use a small MLP with two FC layers (with 128-channel hidden layers) and ReLU activation.

We use the Adam optimizer \cite{adam} with initial learning rates of 0.02 for tensor factors and (when using neural features) 0.001 for the MLP decoder.
We optimize our model for $T$ steps with a batch size of 4096 pixel rays on a single Tesla V100 GPU (16GB).
We apply a feature grid with a total number of $N^3$ voxels; the actual resolution of each dimension is computed based on the shape of the bounding box. 
To achieve coarse-to-fine reconstruction, we start from an initial low-resolution grid with $N_0^3$ voxels with $N_0=128$; we then
 upsample the vectors and matrices linearly and bilinearly at steps 2000, 3000, 4000, 5500, 7000 with the numbers of voxels interpolated between $N_0^3$ and $N^3$ linearly in logarithmic space. 
 Please refer to Sec.~\ref{sec:exp} for the analysis on different total steps ($T$), different resolutions ($N$), and different number of total components ($3R_\Dens+3R_\Rad$).

\section{Experiments}
\label{sec:exp}
We now present an extensive evaluation of our tensorial radiance fields. We first analyze our decomposition techniques, the number of components, grid resolutions, and optimization steps. We then compare our approach with previous and concurrent works on both 360$^\circ$ objects and forward-facing datasets.

\boldstartspace{Analysis of different TensoRF models.}
We evaluate our TensoRF on the Synthetic NeRF dataset \cite{mildenhall2020nerf} using both CP and VM decompositions with different numbers of components and different numbers of grid voxels.
Table~\ref{tab:decomp} shows the averaged rendering PSNRs, reconstruction time, and model size for each model. We use the same MLP decoding function (as described in Sec.~\ref{sec:impl}) for all variants and optimize each model for 30k steps with a batch size of 4096.

Note that both TensoRF-CP and TensoRF-VM achieve consistently better rendering quality with more components or higher grid resolutions. 
TensoRF-CP achieves super compact modeling; even the largest model with 384 components and $500^3$ voxels requires less than 4MB.
This CP model also achieves the best rendering quality in all of our CP variants, leading to a high PSNR of 31.56, which even outperforms vanilla NeRF (see Tab.~\ref{table:results}).

On the other hand, because it compresses more parameters in each component, TensoRF-VM achieves significantly better rendering quality than TensoRF-CP; even the smallest TensoRF-VM model with only 48 components and $200^3$ voxels is able to outperform the best CP model that uses many more components and voxels.
Remarkably, the PSNR of 31.81 from this smallest VM model (which only takes $8.6$ MB) is already higher than the PSNRs of NeRF and many other previous and concurrent techniques (see Tab.~\ref{table:results}). 
In addition, 192 components are generally adequate for TensoRF-VM; doubling the number to 384 only leads to marginal improvement. 
TensoRF-VM with $300^3$ voxels can already lead to high PSNRs close to or greater than 33, while retaining compact model sizes ($<$72MB).
Increasing the resolution further leads to improved quality, but also larger model size.

Also note that all of our TensoRF models can achieve very fast reconstruction. Except for the largest VM model, all models finish reconstruction in less than 30 min, significantly faster than NeRF and many previous methods (see Tab.~\ref{table:results}). 
% For the rest experiments, we use our best CP model and VM models with $300^3$ voxels and components $\leq 192$ to generate most of our results, while many of our models can also generate high-quality and even better results. 
% \begin{table*}[tpb]
%     \centering
%     % \begin{subtable}[t]{\linewidth}
%         \centering
%         \begin{tabular}{c|ccccc|ccccc}
%         \hline
%                         & 5k & 15k & 30k & 60k & 100k       & 5k & 15k & 30k & 60k & 100k\\
%         \hline\hline
%             CP-384 & 28.37 & 30.80 & 31.56 & 32.03 & 32.18  & 03:03 & 11:30 & 25:00 & 51:47 & 85:38\\
%             VM-48  & 29.28 & 31.80 & 32.39 & 32.68 & 32.84  & 01:57 & 06:21 & 13:51 & 27:20 & 45:58\\
%             VM-96  & 29.65 & 32.26 & 32.86 & 33.17 & 33.29  & 02:01 & 06:41 & 14:08 & 28:57 & 48:35\\
%             VM-192 & 29.86 & 32.52 & 33.14 & 33.44 & 33.54  & 02:16 & 08:08 & 17:37 & 35:50 & 60:10\\
%             VM-384 & 29.95 & 32.62 & 33.21 & 33.52 & 33.64  & 02:51 & 11:30 & 25:14 & 52:50 & 89:33\\
%         \hline
%         \end{tabular}
%     \caption{PSNRs and time of CP and VM models with different training steps.}\vspace{-8mm}
%     \label{tab:steps}
% \end{table*}

\boldstartspace{Optimization steps.}
We further evaluate our approach with different optimization steps for our best CP model and the VM models with $300^3$ voxels. PSNRs and reconstruction time are shown in Tab.~\ref{tab:steps}.
All of our models consistently achieve better rendering quality with more steps.
Our compact CP-384 model (3.9MB) can even achieve a PSNR greater than 32 after 60k steps, higher than the PSNRs of all previous methods in Tab.~\ref{table:results}.
On the other hand, our VM models can quickly achieve high rendering quality in very few steps. With only 15k steps, many models achieve high PSNRs that are already state-of-the-art.

% \definecolor{bronze}{rgb}{0.871,0.773,0.286}
 \definecolor{bronze}{rgb}{1,1,0.6}
\definecolor{silve}{rgb}{0.969,0.796,0.600}
% \definecolor{silve}{rgb}{0.839,0.843,0.843}
\definecolor{gold}{rgb}{0.941,0.592,0.600}
% \definecolor{gold}{rgb}{0.996,0.882,0.0039}
% \definecolor{gold}{rgb}{0.946,0.682,0.502}

% \newcommand{\gold}[1]{\colorbox{gold}{\textbf{#1}}}
% \newcommand{\silve}[1]{\colorbox{silve}{\textbf{#1}}}
% \newcommand{\bronze}[1]{\colorbox{bronze}{\textbf{#1}}}
\newcommand{\gold}[1]{\colorbox{gold}{{#1}}}
\newcommand{\silve}[1]{\colorbox{silve}{{#1}}}
\newcommand{\bronze}[1]{\colorbox{bronze}{{#1}}}

\setlength{\tabcolsep}{4pt}
\begin{table}[t]
\centering
\resizebox{\textwidth}{!}{%

\begin{tabular}{l|cc|cc|cc|cc|cc}
& \multicolumn{6}{c|}{Synthetic-NeRF} & \multicolumn{2}{c|}{NSVF} & \multicolumn{2}{c}{TanksTemples} \\
Method & BatchSize & Steps & Time $\downarrow$  & Size(MB)$\downarrow$ & PSNR$\uparrow$ & SSIM$\uparrow$ & PSNR$\uparrow$ & SSIM$\uparrow$ & PSNR$\uparrow$ & SSIM$\uparrow$ \\
\hline

% \midrule
    SRN~\cite{sitzmann2019scene}          &  - & -       & $>$10h     & - &22.26 & 0.846 &   24.33 & 0.882  &  24.10 & 0.847 \\%~\cite{sitzmann2019scene}
    NSVF~\cite{liu2020neural}             & 8192 & 150k  & $>$48$^*$h &  -    & 31.75 & 0.953 &   35.18 & \silve{0.979}   &  \silve{28.48} & 0.901 \\%~\cite{liu2020neural}
    NeRF~\cite{mildenhall2020nerf}        & 4096 & 300k  & $\sim$35h  &  \silve{5.00}    & 31.01 & 0.947 &   30.81 & 0.952   &  25.78 & 0.864 \\%~\cite{mildenhall2020nerf}
    SNeRG~\cite{hedman2021baking}         & 8192 & 250k  & $\sim$15h      & 1771.5& 30.38 & 0.950 &  -      & -       & -      & -     \\%~\cite{hedman2021baking}
    PlenOctrees~\cite{yu2021plenoctrees}  & 1024 & 200k  & $\sim$15h      & 1976.3& 31.71 & \bronze{0.958} &   -     & -       &  27.99 & \silve{0.917} \\%~\cite{yu2021plenoctrees}
    Plenoxels~\cite{yu2021plenoxels}      & 5000 & 128k  & \silve{11.4m}  & 778.1 & 31.71 & \bronze{0.958} &   -     & -       &  27.43 & 0.906 \\%~\cite{yu2021plenoxels}
    DVGO~\cite{sun2021direct}             & {5000} & \silve{30k}   & 15.0m          & 612.1 & 31.95 & 0.957 &   35.08 & 0.975.  &  \bronze{28.41} & 0.911 \\%~\cite{sun2021direct}
\hline
    Ours-CP-384                            & {4096} & \silve{30k}  & 25.2m           &\gold{3.9}    & 31.56 & 0.949 &   34.48 & 0.971   &  27.59 & 0.897 \\
    Our-VM-192-SH                          & {4096} & \silve{30k}  & 16.8m           &71.9   & 32.00 & 0.955 &   35.30 & 0.977   &  27.81 & 0.907 \\
    Ours-VM-48                             & {4096} & \silve{30k}  & \bronze{13.8m}  &\bronze{18.9}   & \bronze{32.39} & 0.957 &   \silve{35.34} & 0.976   &  28.06 & 0.909 \\
    Ours-VM-192                            & {4096} & \gold{15k}  & \gold{8.1m}     &71.8   & \silve{32.52} & \silve{0.959} &   \bronze{35.59} & \bronze{0.978}   &  28.07 & \bronze{0.913} \\
    Ours-VM-192                            & {4096} & \silve{30k}  & 17.4m     &71.8   & \gold{33.14} & \gold{0.963} &   \gold{36.52} & \gold{0.982}   &  \gold{28.56} & \gold{0.920} \\
    
% \hline
\end{tabular}
} 
\vspace{1mm}
\caption{We compare our method with previous and concurrent novel view synthesis methods on three datasets. All scores of the baseline methods are directly taken from their papers whenever available. We also report the averaged reconstruction time and model size for the Synthetic-NeRF dataset. NVSF requires 8 GPUs for optimization (marked by a star), while others run on a single GPU. DVGO's 30k steps correspond to 10k for coarse and 20k for fine reconstruction.} \vspace{-6mm}

\label{table:results}  
\end{table}
\setlength{\tabcolsep}{1.4pt}

\begin{table*}[tpb]
    \centering
    % \begin{subtable}[t]{\linewidth}
        \centering
        \renewcommand\tabcolsep{5.0pt}
        \begin{tabular}{c|c|c|c|c}
        \hline
        &$\#$Comp                        &         $200^3$       &    $300^3$            &      $500^3$\\
        \hline\hline
    \multirow{4}{*}{TensoRF-CP} &48   & 27.98/09:29/0.74      &  28.24/11:45/1.09     &  28.38/14:20/1.85 \\
                                &96   & 28.50/09:57/0.88      &  28.83/12:12/1.29     &  29.06/15:27/2.18 \\
                                &192  & 29.50/11:09/1.08      &  29.99/13:41/1.59     &  30.33/18.03/2.66 \\
                                &384  & 30.47/14:41/1.59      &  31.08/18:09/2.33     &  31.56/25:11/3.93 \\

        \hline
    \multirow{4}{*}{TensoRF-VM} &48   & 31.81/11:29/08.6 &     32.39/13:51/23.5 &   32.63/18:17/55.8  \\
                                &96   & 32.33/11:54/16.5 &     32.86/14:08/37.3 &   33.06/20:11/105. \\
                                &192  & 32.63/13:26/32.3 &     33.14/17:36/76.7 &   33.31/27.18/204. \\
                                &384  & 32.69/17:24/63.4 &     33.21/25:14/143. &   33.39/43:19/397.\\
        \hline
        \end{tabular}
    \caption{
    We compare the averaged PSNRs / optimization time (mm:ss) / model sizes (MB) of CP and VM TensoRF models on Synthetic NeRF dataset \cite{mildenhall2020nerf} with different numbers of components and grid resolutions, optimized for 30k steps.
    %  We optimize TensoRF with both CP decomposition and VM decomposition with different number of total components for 30k steps and compare the rendering quality (PSNR) / optimization time (mm:ss) / model size (MB).
    }\vspace{-6mm}
    \label{tab:decomp}
\end{table*}

\begin{table*}[!t]

  \begin{minipage}{0.51\columnwidth}
    \centering
    % \hrule
    \resizebox{\textwidth}{!}{%
        \begin{tabular}{c|cccc|cccc}
        % \hline
                        & 5k & 15k & 30k & 60k        & 5k & 15k & 30k & 60k\\
        \hline%\hline
            CP-384 & 28.37 & 30.80 & 31.56 & 32.03  & 03:03 & 11:30 & 25:11 & 51:47\\
            VM-48  & 29.28 & 31.80 & 32.39 & 32.68  & 01:57 & 06:21 & 13:51 & 27:20\\
            VM-96  & 29.65 & 32.26 & 32.86 & 33.17  & 02:01 & 06:41 & 14:08 & 28:57\\
            VM-192 & 29.86 & 32.52 & 33.14 & 33.44  & 02:16 & 08:08 & 17:37 & 35:50\\
            VM-384 & 29.95 & 32.62 & 33.21 & 33.52  & 02:51 & 11:30 & 25:14 & 52:50\\
        \hline
        \end{tabular}
    }
    \caption{PSNRs and time of CP and VM models with different training steps on the Synthetic-NeRF dataset \cite{mildenhall2020nerf}.}
    \label{tab:steps}
  \end{minipage}\hfill % maximize the horizontal separation
    \begin{minipage}{0.46\columnwidth}
    \centering
    % \hrule
    \resizebox{\textwidth}{!}{%
        \begin{tabular}{l|cccc}
        % \hline
        %\multicolumn{5}{c}{\textbf{Forward-facing}}  \\
        % \hline
        Method  & Time $\downarrow$ & Size & PSNR$\uparrow$ & SSIM$\uparrow$\\
        \hline
        
        % \midrule
            NeRF~\cite{mildenhall2020nerf}        & 36h          & \textbf{5.00M}    & 26.50 & 0.811\\%~\cite{mildenhall2020nerf}
            Plenoxels~\cite{yu2021plenoxels}      & 24:00m       & 2.59G & 26.29 & 0.829 \\%~\cite{yu2021plenoxels}
            % DVGO~\cite{sun2021direct}             & 15.0m        & 612.1 & 31.95 & 0.957  \\%~\cite{sun2021direct}
        \hline
            % Our-CP-384                          & 25.2min       &3.9    & 31.56 & 0.949 \\
            Ours-VM-48                             & \textbf{19:44m}        &90.4M   & 26.51 & 0.832\\
            Ours-VM-96                            & 25.43m        &179.7M   & \textbf{26.73} & \textbf{0.839} \\
        \hline
    \end{tabular}
    }
    \caption{Quantitative comparisons of our method with NeRF and Plenoxels on forward-facing scenes \cite{llff}.}
    \label{table:score_llff} 
  \end{minipage}\vspace{-10mm}
\end{table*}

\begin{figure}[t]
    \centering
    \includegraphics[width=0.95\linewidth]{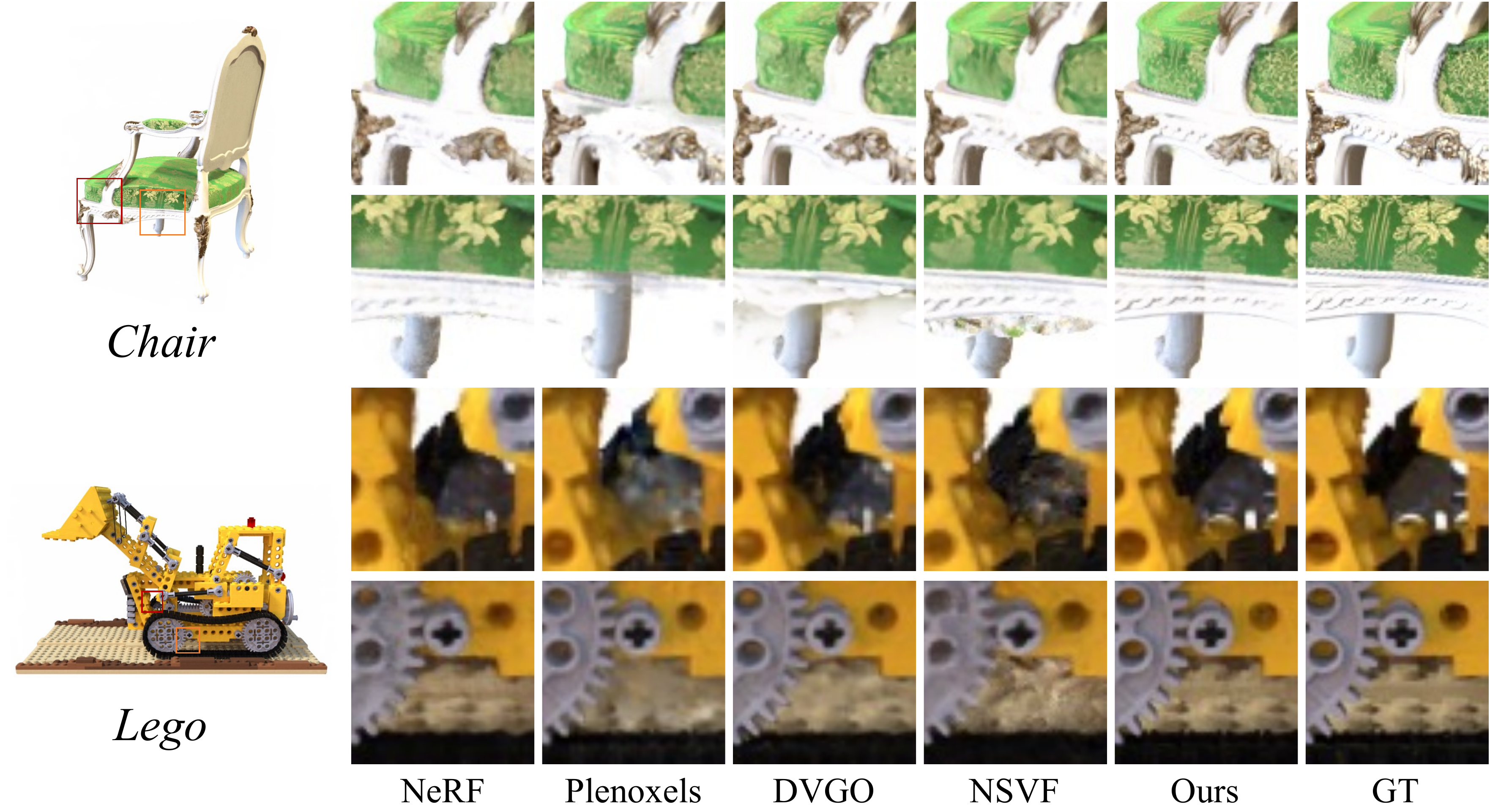}
    \vspace{-2mm}
    \caption{Qualitative results of our VM-192-30k model and comparison methods (NeRF\cite{mildenhall2020nerf}, plenoxels \cite{yu2021plenoxels}, DVGO \cite{sun2021direct}, NSVF \cite{liu2020neural}) on two Synthetic NeRF scenes. }
    \label{fig:restuls}\vspace{-5mm}
\end{figure}

\boldstartspace{Comparisons on 360$^\circ$ scenes.}
We compare our approach with state-of-the-art novel view synthesis methods, including previous works (SRN\cite{sitzmann2019scene}, NeRF\cite{mildenhall2020nerf}, NSVF\cite{liu2020neural}), SNeRG\cite{hedman2021baking}, PlenOctrees\cite{yu2021plenoctrees}) and concurrent works (Plenoxels \cite{yu2021plenoxels}, DVGO\cite{sun2021direct}).
In particular, we compare with them using our best CP model and our VM models (300$^3$ voxels) with 48 and 192 components.
We also show a 192-component VM model with spherical harmonics shading function.
Table ~\ref{table:results} shows the quantitative results (PSNRs and SSIMs) of ours and comparison methods on three challenging datasets, where we also show the corresponding batch size, optimization steps, time, and final output model size for each model, to compare all methods from multiple perspectives. 
Note that all of our CP and VM models can outperform NeRF on all three datasets while taking substantially less optimization time and fewer steps.
Our best VM-192 model can even achieve the best PSNRs and SSIMs on all datasets, significantly outperforming the comparison methods.
Our approach can also achieve qualitatively better renderings with more appearance and geometry details and less outliers, as shown in 
Fig.~\ref{fig:restuls}.

% Note that our VM models (both SH and MLP) can achieve the state-of-the art rendering quality on all three datasets.
% In particular, our results are consistently better than SRN, NeRF and DVGO on all three datasets.
% SNeRG and Plenoxels only provides results on the Synthetic-NeRF datasets, and our approach outperforms both on this dataset.
% Compared to PlenOctress and NSVF, our approach outperforms them on both NeRF-Synthetic and NSVF-Synthetic datasets, while achieving comparable quality on the Tanks and Temples dataset (with higher SSIM/PSNR and lower PSNR/SSIM).
% Example rendering results with comparisons are shown in Fig.xxx, which clearly demonstrates the high rendering quality of our tensorial approach.

% Our high rendering quality is achieved with high efficiency in terms of both model compactness and optimization time.
% Our high rendering quality is achieved with high efficiency.
Our models are highly efficient, which all require less than 75MB space and can be reconstructed in less than 30 min.
% On the contrary, though very compact, NeRF requires about 1.5 days for optimization.
% % All of our models attain high-quality reconstruction in less than 30 min, leading to more than 70x speed up compared to NeRF.
% Our models (all taking less than 30 min) lead to more than 70x speed up compared to NeRF.
This corresponds to more than 70x speed up compared to NeRF that requires about 1.5 days for optimization.
Our CP model is even more compact than NeRF.
Moreover, SNeRG and PlenOctrees require pre-training a NeRF-like MLP, requiring long reconstruction time too.
% ; their methods rely on building explicit sparse features grids, leading to substantially larger model sizes
DVGO and Plenoxels are concurrent works, which can also achieve fast reconstruction in less than 15 min. 
However, as both are voxel-based methods and directly optimize voxel values, they lead to huge model sizes 
% in the order of GBs of memory 
similar to previous voxel-based methods like SNeRG and PlenOctrees.
In contrast, we factorize feature grids and model them with compact vectors and matrices, leading to substantially smaller model sizes.
% In contrast, we leverage tensor decomposition techniques to factorize feature grids and model them with compact vectors and matrices, leading to a substantially smaller model size than these previous and concurrent grid-based methods.
Meanwhile, our VM-192 can even reconstruct faster than DVGO and Plenoxels, taking only 15k steps, and achieving better quality in most cases. 
In fact, Plenoxels' fast reconstruction relies on quickly optimizing significantly more steps ($>4$ times our steps) with their CUDA implementation. 
Our models are implemented with standard PyTorch modules and already achieve much better rendering quality with fewer steps taking comparable and even less reconstruction time than Plenoxels. Note that our SH model essentially represents the same underlying feature grid as Plenoxels but can still lead to more compact modeling and better quality with fewer steps, showing the advantages of our factorization based modeling.
In general, our approach enables fast reconstruction, compact modeling, and photo-realistic rendering simultaneously.
% Besides, our VM-192 can also reconstruct faster than DVGO and achieve better quality on two of the three datasets. 
% On the other hand, Plenoxels' fast reconstruction relies on quickly optimizing significantly more steps ($>3$ times our steps) with their customized cuda kernals. 
% Our results are instead generated with standard PyTorch implementation and cannot take many steps in the same period.
% Nonetheless, our VM models achieve much better rendering quality with fewer steps and comparable reconstruction time; this even includes our SH model that essentially represents the same underlying feature grid as Plenoxels, which clearly shows the benefits of our factorization based model.
% In general, our approach enables fast reconstruction, compact modeling, and photo-realistic rendering at the same time.

\boldstartspace{Forward-facing scenes.}
Our approach can also achieve efficient and high-quality radiance field reconstruction for forward-facing scenes.
We show quantitative results of our two VM models on the LLFF dataset \cite{llff} and compare with NeRF and Plenoxels in Tab.~\ref{table:score_llff}.
Our models outperform the previous state-of-the-art NeRF and take significantly less reconstruction time.
Compared with Plenoxels \cite{yu2021plenoxels}, our approach leads to comparable or faster reconstruction speed, better quality, and substantially smaller model sizes.
% \fv{Our approach can achieve high rendering quality, comparable to the state-of-the-art NeRF, while taking significantly less reconstruction time.}

\section{Conclusion.}
We have presented a novel approach for high-quality scene reconstruction and rendering. We propose a novel scene representation -- TensoRF which leverages tensor decomposition techniques to model \fv{and reconstruct} radiance fields compactly as factorized low-rank tensor components.
While our framework accommodates classical tensor factorization techniques (like CP), we introduce a novel vector-matrix decomposition that leads to better reconstruction quality and faster optimization speed.
Our approach enables highly efficient radiance field reconstruction in less than 30 min per scene, leading to better rendering quality compared to NeRF that requires substantially longer training time (20+ hours).
Moreover, our tensor factorization based method achieves high compactness, leading to a memory footprint of less than 75MB, substantially smaller than many other previous and concurrent voxel-grid based methods. 
\fv{We hope our findings in tensorized low-rank feature modeling can inspire other modeling and reconstruction tasks.}

\bibliographystyle{splncs04}
\bibliography{tex/reference}

\appendix
% \section{Correction.}
% We correct two issues in the main paper that we found after submission. We apologize for the inconvenience.

% \begin{itemize}
%     \item In Tab.~\ref{table:results}, the correct PSNR/SSIM of Ours-CP-384 is 34.48/0.971. All other numbers are correct. Please refer to Sec.~\ref{sec:perscene} for per-scene breakdown.
%     \item In Fig.~\ref{fig:restuls}, the (left-most) two small crops of NeRF of the Chair scene are incorrect. We show the correct ones in Fig.~\ref{fig:restuls_correct}.
% \end{itemize}

% \begin{figure}[t]
%     \centering
%     \includegraphics[width=0.95\linewidth]{figs/comp_syn_nerf.pdf}
%     \vspace{-2mm}
%     \caption{Qualitative results of our VM-192-30k model and comparison methods (NeRF\cite{mildenhall2020nerf}, plenoxels \cite{yu2021plenoxels}, DVGO \cite{sun2021direct}, NSVF \cite{liu2020neural}) on two Synthetic NeRF scenes. }
%     \label{fig:restuls_correct}\vspace{-5mm}
% \end{figure}

\renewcommand{\gold}[1]{\colorbox{bronze}{{#1}}}
\renewcommand{\gold}[1]{\textbf{#1}}

\section{TensoRF Representation Details.}
\begin{figure}[t]
    \centering
    \includegraphics[width=\linewidth]{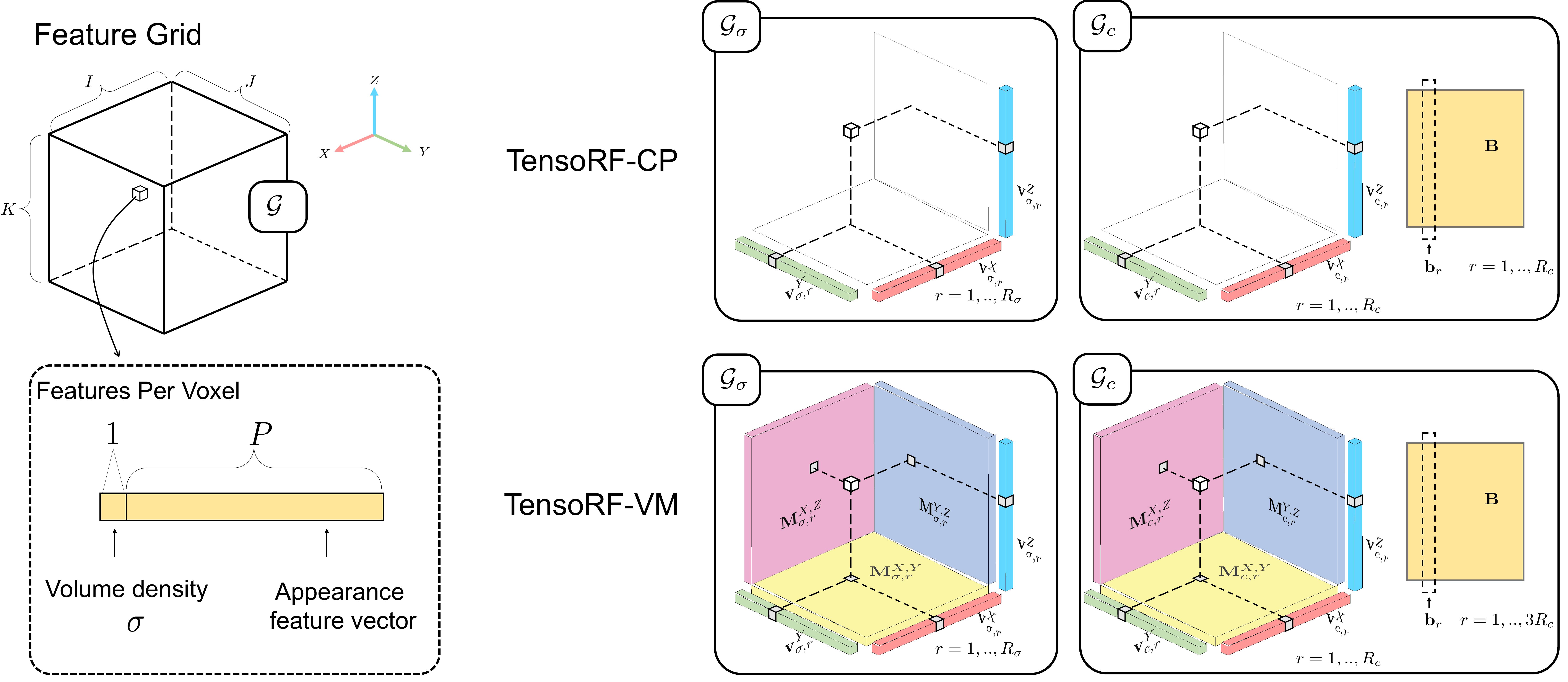}
    % \captionof{figure}{\jet{caption}}
    \caption{Feature grids and factorized tensors in TensoRF. We leverage a regular voxel grid $\Grid$, covering a 3D scene, to model a radiance field of the scene. Each voxel of $\Grid$ contains multi-channel features, where one channel represents the volume density ($\Dens$) and the remaining $P$ channels lead to  an appearance feature vector ($f_\Rad$) for computing view-dependent colors. We split the density and appearance features into two feature grids $\Grid_\Dens$ and $\Grid_\Rad$, consider them as 3D and 4D tensors, and factorize them into compact factors with outer products. TensoRF with CP decomposition factorize the tensors into only vectors and TensoRF with VM decomposition factorize the tensor into vector and matrix factors (Eqn.~\ref{eqn:dense-vmr}, \ref{eqn:rad-vmr}). Note that, each voxel of the original grid is only related to one value from each XYZ-mode vector/matrix factor in both decompositions, which are marked in the figure.}
    \label{fig:grids}
\end{figure}

We illustrate the feature grid of our tensorial radiance field and the tensor factors in TensoRF with both CP and VM decompositions in Fig.~\ref{fig:grids}.

\boldstartspace{Number of components.}
The total number of tensor components ($\#$Comp) is $(R_\Dens+R_\Rad)$ for TensoRF-CP and $3(R_\Dens+R_\Rad)$ for TensoRF-VM (because VM has three types of components). Therefore, the $R$ we use for TensoRF-CP is three times as large as the $R$ used for TensoRF-VM to achieve the same number of components shown in Tab.~2.
We also find that using $R_\Dens<R_\Rad$ is usually better than $R_\Dens=R_\Rad$ when $R_\Dens$ is large enough ($>8$).
In particular, for TensoRF-VM, we use $R_\Dens=R_\Rad=8$ for $\#\text{Comp}=48$; $R_\Dens=8, R_\Rad=24$ for $\#\text{Comp}=96$; $R_\Dens=16,R_\Rad=48$ for $\#\text{Comp}=192$; $R_\Dens=32,R_\Rad=96$ for $\#\text{Comp}=384$. Note that, as discussed in Eqn.~\ref{eqn:vmrrr},\ref{eqn:vmr}, here we apply the same number of components for $\Comp^X$, $\Comp^Y$, $\Comp^Z$ with $R_1=R_2=R_2=R$ for both density and appearance (where $R$ is $R_\Dens$ and $R_\Rad$ respectively), assuming the three spatial dimensions are equally complex.

\boldstartspace{Forward-facing settings.}
We use the above settings with $R_1=R_2=R_2$, for all 360$^\circ$ object datasets in Tab.~\ref{table:results}.  On the other hand, Forward-facing scenes apparently appear differently in the three dimensions; especially, in NDC space, the $X$ and $Y$ spatial modes (corresponding to the image plane) contains more appearance information that is visible to rendering viewpoints. We therefore use more components for the $X-Y$ plane, corresponding to $\Comp^Z=\Vector^Z \circ \Matrix^{X,Y}$. In this case, these $\Comp^Z$ components can also be seen as special compressed versions of neural MPIs. In particular, the detailed numbers of components we use for generating the results in Tab.~\ref{table:score_llff} are: for $\#$Comp=48 $R_{\Dens,1}=R_{\Dens,2}=4$,$R_{\Dens,2}=16$,$R_{\Rad,1}=R_{\Rad,2}=4$,$R_{\Rad,2}=16$; for $\#$Comp=96, $R_{\Dens,1}=R_{\Dens,2}=4$,$R_{\Dens,2}=16$,$R_{\Rad,1}=R_{\Rad,2}=16$,$R_{\Rad,48}=16$.

\boldstartspace{Number of parameters.}
We briefly discuss the number of parameters in our model.
With the same $\#$Comp, when $I=J=K$ and $R_\Dens+R_\Rad=R$, the total number of parameters used for TensoRF-CP is $3KR+PR_\Rad$; for Tensor-VM, the number is $K^2R+KR+PR_\Rad$ (here considering $R_\Dens/3$, $R_\Rad/3$ are used to make the $\#$Comp the same as TensoRF-CP ). For example, for a $300\times300\times300$ feature grid with $P=27$ channels (plus one density channel), the total number of parameters in a dense grid is 756 M; the number of parameters used for TensoRF-CP (when $R=192$) is 360 K; the number of parameters used for TensoRF-VM (when $R=192$) is 17 M.
Our CP and VM model can achieve $0.048\%$ and $2.25\%$ compression rates respectively.

% \boldstartspace{Trilinear tensor index.}
% We illustrate the linear/bilinear interpolation in factors is identical to trilinearly indexing in tensor:

% \begin{align}
%     \dddot{\Lambda}(\Tensor) &= \sum_{r=1}^R \dot{\Lambda}(\Vector_r^{1}) \OuterP \dot{\Lambda}(\Vector_r^{2}) \OuterP \dot{\Lambda}(\Vector_r^{3})\\
%     &= \sum_{r=1}^{R_1} \dot{\Lambda}(\Vector_r^{1}) \OuterP \ddot{\Lambda}(\Matrix_r^{2,3}) + \sum_{r=1}^{R_2}\dot{\Lambda}(\Vector_r^{2}) \OuterP \ddot{\Lambda}(\Matrix_r^{1,3}) + \sum_{r=1}^{R_3}\dot{\Lambda}(\Vector_r^{3}) \OuterP \ddot{\Lambda}(\Matrix_r^{1,2}) 
%     \label{eqn:tri-index}
% \end{align}

% Where $ \dot{\Lambda}, \ddot{\Lambda}, \dddot{\Lambda}$ are the linear/bilinear/trilinear interpolation respectively.

\section{More Implementation Details.}

\boldstartspace{Loss functions.}
As described in sec. \ref{sec:rendering_reconstruction}, we apply a L2 rendering loss and additional regularization terms to optimize our tensor factors for radiance field reconstruction.
In general, this loss function is expressed by
% To optimize the matrix/vectors of our TensoRF, we randomly sample rays from the given multi-view input images and compute the ray color by sampling shading points along the rays as descript in sec. \ref{sec:rendering_reconstruction}. we then evaluate the min-batch loss with an L2 rendering loss and one additional regularization term:

\begin{equation}
\begin{aligned}
    \loss = \|\Color - \tilde{\Color} \|^2_2 + \omega  \cdot \loss_{reg}
\label{eq:loss}
\end{aligned}
\end{equation}
Where $\tilde{\Color}$ is the groundtruth color, $\omega$ is weight of the regularization term.

To encourage the sparsity in the parameters of our tensor factors, we apply the standard L1 regularization, which we find is effective in improving the quality in extrapolating views and removing floaters/outerliers in final renderings. Note that, unlike previous methods \cite{hedman2021baking,sun2021direct} that penalize predicted per-point density with a Cauchy loss or entropy loss, our L1 regularizer is much simpler and directly applied on the parameters of tensor factors. 
We find that it is sufficient to  apply the L1 sparsity loss only on the density parameters, expressed by
% which is a stander regularization term in signal processing and machine learning:
\begin{equation}
\begin{aligned}
    \loss_{L1} = \frac{1}{N}\sum_{r=1}^{R_\Dens} (\|\Matrix_{\Dens,r}\| + \|\Vector_{\Dens, r}\|),
\label{eq:loss_reg}
\end{aligned}
\end{equation}
where $\|\Matrix_{\Dens,r}\|$ and $\|\Vector_{\Dens, r}\|)$ are simply the sum of absolute values of all elements, and $N= R_\Dens \cdot (I \cdot J + I \cdot K+J \cdot K + I + J + K)$ is the total number of parameters.
We use this L1 sparsity loss with a $\omega=0.0004$ for the Synthetic NeRF and Synthetic NSVF datasets. An ablation study on this L1 loss on the Synthetic NeRF dataset is shown in Tab.~\label{tab:ab_regularization}.

For real datasets that have very few input images (like LLFF\cite{llff}) or imperfect capture conditions (like Tanks and Temples \cite{Knapitsch2017} that has varying exposure and inconsistent masks), we find a TV loss is more efficient
than the L1 sparsity loss, expressed by

\begin{equation}
\begin{aligned}
    \loss_{TV} =  \frac{1}{N} \sum (\sqrt{\triangle^2 \Comp^m_{\Dens,r}} + 0.1 \cdot \sqrt{\triangle^2 \Comp^m_{\Color,r}}),
\label{eq:loss_reg}
\end{aligned}
\end{equation}

Here $\triangle^2$ is the squared difference between the neighboring values in the matrix/vector factors; we apply a smaller weight (weighted by 0.1 additionally) on appearance parameters in the TV loss.
We use $\omega=1$ when using this TV loss.

\boldstart{Binary occupancy volume.}
To facilitate reconstruction, we compute a binary occupancy mask grid at steps 2000 and 4000 using the volume density prediction from the intermediate TensoRF model to avoid computation in empty space.
%  Our method requires a bounding box to determine the space of the underlying tensor feature grid.
For datasets that do not provide bounding boxes, we start from a conservatively large box and leverage the occupancy mask computed at step 2000 to re-compute a more compact bounding box, with which we shrink and resample our tensor factors, leading to more precise modeling. For forward-facing scenes in the LLFF dataset \cite{llff}, we apply NDC transformation that bounds the scene in a perspective frustum.

\boldstart{More details.}
As described in Sec.~\ref{sec:impl}, we use a small two-layer MLP with 128 channels in hidden layers as our neural decoding function.
In particular, the input to this MLP contains the viewing direction and the features recovered by our tensor factors (no xyz positions are used). Similar to NeRF and NSVF \cite{mildenhall2020nerf,liu2020neural}, we also apply frequency encodings (with Sin and Cos functions) on both the viewing direction and features. Unlike NeRF that uses ten different frequencies, we use only two.

During optimization, we also apply an exponential learning rate decay to make the optimization more stable when the reconstruction is being finished. Specifically, we decay our initial learning rates at every training step, until decayed by a factor of 0.1 in the end of the optimization.

% Note that, NeRF \cite{martin2021nerf} reguralize the optimization process through globally parameterized architecture (i.e., MLP) and Fourier space transformation (i.e., position encoding), which naturally share information and regularize across points. However, this results in computer complexity and time consumer.

\section{More Evaluation.}
We perform an ablation study to evaluate our L1 regularization.
% We perform an ablation study of our CP and VM factorisation to understand the role of each component. 
Tab. \ref{tab:ab_regularization} shows how our framework performs by removing the L1 regularization on the Synthetic-NeRF dataset, our models exceed NeRF fidelity ($31.01$ in average) even without regularization. 
We observe the performance gap between w/ and w/o L1 regularization is mostly caused by the floaters in the empty space.
% which resulting in our shrinking process couldn't obtain a tight bounding box, thus the trained  basis include a large empty space. 
We also provide more results on our model with different numbers of training steps in Tab.~\ref{tab:steps_more}, which is basically a detailed version of Tab.~\ref{tab:steps} with more settings.
These results showcase that our models consistently improve when training with more iterations.

% In Tab. 
% \ref{steps_more}, we iteratively increase the number of iterations and basis, this ablation sheds light on larger basis number or iteration steps is able to uniformly boost our performance.

\begin{table*}[htpb]
    \centering
    % \begin{subtable}[t]{\linewidth}
        \centering
        \renewcommand\tabcolsep{5.0pt}
        \begin{tabular}{c|cccc}
        % \hline
                        & PSNR & SSIM & LPIPS$_{VGG}$&LPIPS$_{Alex}$ \\
        \hline%\hline
            CP-384      & 31.56/31.23 & 0.949/0.947 & 0.076/0.078 & 0.041/0.043 \\
            VM-48       & 32.39/31.71 & 0.957/0.953 & 0.057/0.062 & 0.032/0.036  \\
            VM-192-SH   & 32.00/31.14 & 0.955/0.949 & 0.058/0.068 & 0.058/0.044  \\
            VM-192      & 33.14/32.43 & 0.963/0.960 & 0.047/0.052 & 0.027/0.030  \\
        \hline
        \end{tabular}
    \caption{We compare the averaged scores against w/o L1 regularization on the Synthetic-NeRF dataset.}
    \label{tab:ab_regularization} \vspace{-2mm}
\end{table*}

\begin{table*}[htpb]
    \centering
    % \begin{subtable}[t]{\linewidth}
        \centering
        \renewcommand\tabcolsep{4.0pt}
        \begin{tabular}{c|ccccccccccc}
        \hline
                        & 5k & 8k & 10k & 12k & 15k & 20k & 30k & 40k & 60k & 100k \\
        \hline\hline
            CP-192 & 28.38 & 29.13 & 29.93 & 30.38 & 30.80 & 31.18 & 31.56 & 31.75 & 32.03 & 32.18 \\
            VM-48  & 29.28 & 30.39 & 31.11 & 31.47 & 31.80 & 32.08 & 32.39 & 32.55 & 32.68 & 32.84 \\
            VM-96  & 29.65 & 30.72 & 31.52 & 31.93 & 32.26 & 32.56 & 32.86 & 33.00 & 33.17 & 33.29 \\
            VM-192 & 29.86 & 30.93 & 31.74 & 32.17 & 32.52 & 32.85 & 33.14 & 33.27 & 33.44 & 33.54 \\
            VM-384 & 29.95 & 30.88 & 31.75 & 32.20 & 32.62 & 32.94 & 33.21 & 33.35 & 33.52 & 33.64 \\
        \hline
        \end{tabular}
    \caption{PSNRs on the Synthetic NeRF datasets with different numbers of training steps. This is more detailed version than Tab.~\ref{tab:steps}.}
    \label{tab:steps_more} \vspace{-4mm}
\end{table*} 

% \paragraph{SH vs neural.}
% We can also see that Ours-neural is consistently better than Ours-SH. This is because an MLP function can express more complex functions. Note that, we already assign Ours-SH with a higher resolution grid (about 1.5x higher per dimension), but the limited representationality of the fixed SH function still lead to lower rendering quality.
% On the other hand, Ours-neural with an MLP-based shading function can lead to a much more compact model with higher rendering quality.  

\section{Discussion}

In fact, the reconstruction problem with dense feature grid representation is relatively over-parameterized/under-determined; % and easily suffers from over-fitting
e.g., a $300^3$ grid with $27$ channels has $>$700M parameters, while one hundred $800\times800$ images provide only 64M pixels for supervision.
Therefore, many design choices -- including pruning empty voxels, coarse-to-fine reconstruction, and adding additional losses, which have been similarly used in TensoRF and concurrent works (DVGO, Plenoxels) -- are all essentially trying to reduce/constrain the parameter space and avoid over-fitting. 
In general, low-rank regularization is crucial in addressing many reconstruction problems, like matrix completion \cite{candes2010matrix}, compressive sensing \cite{dong2014compressive}, denoising \cite{ji2010robust};
tensor decomposition has also been widely used in tensor completion \cite{liu2012tensor,gandy2011tensor}, which is similar to our task. 
Tensor decomposition naturally provides low-rank constraints and reduces parameters; this similarly benefits the radiance field reconstruction as demonstrated by our work.

Moreover, TensoRF represents a 5D radiance field function that expresses both scene geometry and appearance;
hence, we believe our 4D tensor is generally low-rank, because a 3D scene typically contains a lot of similar geometry structures and material properties across different locations. 
Note that, in various appearance acquisition tasks, similar low-rank constraints have been successfully applied for reconstructing other functions, including the 4D light transport function in relighting \cite{wang2009kernel} and the 6D SVBRDF function in material reconstruction \cite{zhou2016sparse,nam2018practical} (where a common idea is to model a sparse set of basis BRDFs; this is similar to our modeling of vector components in the feature dimension in the matrix $\textbf{B}$).
We combine low-rank constraints and neural networks from a novel perspective, in tensor-based radiance field reconstruction.
TensoRF essentially models the scene with global basis components, discovering the scene geometry and appearance commonalities across the spatial and feature dimensions.
% We hope our findings in tensorized low-rank feature modeling can inspire other modeling and reconstruction tasks.

\section{Limitations and Future Work.}
Our approach achieves high-quality radiance field reconstruction for 360$^\circ$ objects and forward-facing scenes; however, our method currently only supports bounded scenes with a single bounding box and cannot handle unbounded scenes with both foreground and background content. 
Combining our method with techniques like NeRF++ \cite{zhang2020nerf++} to separately model a foreground field inside a regular box and a background field inside another box defined in a spherical coordinate space can potentially extend our method to address unbounded scenes.
% It is interesting to extend to multiply type of bounding box (e.g., sphere) that enabling unbound scene reconstruction. 
Despite the success in per-scene optimization shown in this paper, 
an interesting future direction is to discover and learn general basis factors across scenes on a large-scale dataset, leveraging data priors to further improve the quality or enable other applications like GANs (as done in GRAF \cite{schwarz2020graf}, GIRRAF \cite{niemeyer2021giraffe} and EG3D \cite{Chan2022EG3D}).

% our method should extend naturally to learn a general basis of modeling geometric and appearance from a large scale date, like fast and compact geometric representation as in Instant-NGP \cite{muller2022instant}, 3D generative models for high resolution reconstruction and rendering.

\section{Acknowledgements}
We would like to thank Yannick Hold-Geoffroy for his useful tips in video animation, Qiangeng Xu for providing some baseline results, and, Katja Schwarz and Michael Niemeyer for providing helpful video materials. This project was supported by NSF grant IIS-1764078 and gift money from Kingstar. \fv{We would also like to thank all anonymous reviewers for their encouraging  comments.}

\section{Per-scene Breakdown.}
\label{sec:perscene}

Tab. \ref{tab:supp_breakdown_nerf}-\ref{tab:supp_breakdown_llff} provide a per-scene break down for quantity metrics in Synthesis-nerf \cite{martin2021nerf}, Synthe-nsvf \cite{liu2020neural}, Tanks$\&$Templates \cite{Knapitsch2017} and forward-facing \cite{llff} dataset.

\begin{table*}[htpb]
    \vspace{2em}
    \centering
    \begin{tabular}{l|c|cccccccc}
    \hline
    Methods & Avg. & {\it Chair} & {\it Drums} & {\it Ficus} & {\it Hotdog} & {\it Lego} & {\it Materials} & {\it Mic} & {\it Ship} \\
    \hline\hline
    \multicolumn{9}{@{}l}{\rule{0pt}{3ex}\bf PSNR$\uparrow$} \\
    \hline
    SRN~\cite{sitzmann2019scene} & 22.26 & 26.96 & 17.18 & 20.73 & 26.81 & 20.85 & 18.09 & 26.85 & 20.60 \\
    NeRF~\cite{mildenhall2020nerf} & 31.01 & 33.00 & 25.01 & 30.13 & 36.18 & 32.54 & 29.62 & 32.91 & 28.65 \\
    NSVF~\cite{liu2020neural} & 31.75 & 33.19 & 25.18 & 31.23 & 37.14 & 32.29 & 32.68 & 34.27 & 27.93 \\
    SNeRG~\cite{hedman2021baking} & 30.38 & 33.24 & 24.57 & 29.32 & 34.33 & 33.82 & 27.21 & 32.60 & 27.97 \\
    PlenOctrees~\cite{yu2021plenoctrees} & 31.71 & 34.66 & 25.31 & 30.79 & 36.79 & 32.95 & 29.76 & 33.97 & 29.42 \\
    Plenoxels~\cite{yu2021plenoxels} & 31.71 & 33.98 & 25.35 & 31.83 & 36.43 & 34.10 & 29.14 & 33.26 & 29.62 \\
    DVGO~\cite{sun2021direct} & 31.95 & 34.09 & 25.44 & 32.78 & 36.74 & 34.64 & 29.57 & 33.20 & 29.13 \\

    \hline
    Ours-CP-384  & 31.56 & 33.60 & 25.17 & 30.72 & 36.24 & 34.05 & 30.10 & 33.77 & 28.84 \\
    Ours-VM-192-SH  & 32.00 & 34.68 & 25.37 & 32.30 & 36.30 & 35.42 & 29.30 & 33.21 & 29.46 \\
    Ours-VM-48  & 32.39 & 34.68 & 25.58 & 33.37 & 36.81 & 35.51 & 29.45 & 33.59 & 30.12 \\
    Ours-VM-192-15k & 32.52 & 34.95 & 25.63 & 33.46 & 36.85 & 35.78 & 29.78 & 33.69 & 30.04\\
    Ours-VM-192-30k & \gold{33.14} & \gold{35.76} & \gold{26.01} & \gold{33.99} & \gold{37.41} & \gold{36.46} & \gold{30.12} & \gold{34.61} & \gold{30.77} \\

    \hline
    \end{tabular}
    \caption{PSNR results on each scene from the {\bf Synthetic-NeRF}~\cite{mildenhall2020nerf} dataset. %Ours $^\dag$ refers to our tensorf representation with direct optimization (i.e., without network) while ours are trained with a small MLP network.
    }
    \label{tab:supp_breakdown_nerf}
    \vspace{2em}
\end{table*}

\begin{table*}[htpb]
    \vspace{2em}
    \centering
    \begin{tabular}{l|c|cccccccc}
    \hline
    Methods & Avg. & {\it Chair} & {\it Drums} & {\it Ficus} & {\it Hotdog} & {\it Lego} & {\it Materials} & {\it Mic} & {\it Ship} \\
    \hline\hline
    % \multicolumn{9}{@{}l}{\rule{0pt}{3ex}\bf PSNR$\uparrow$} \\
    % \hline
    % SRN~\cite{sitzmann2019scene} & 22.26 & 26.96 & 17.18 & 20.73 & 26.81 & 20.85 & 18.09 & 26.85 & 20.60 \\
    % NeRF~\cite{mildenhall2020nerf} & 31.01 & 33.00 & 25.01 & 30.13 & 36.18 & 32.54 & 29.62 & 32.91 & 28.65 \\
    % NSVF~\cite{liu2020neural} & 31.75 & 33.19 & 25.18 & 31.23 & 37.14 & 32.29 & 32.68 & 34.27 & 27.93 \\
    % SNeRG~\cite{hedman2021baking} & 30.38 & 33.24 & 24.57 & 29.32 & 34.33 & 33.82 & 27.21 & 32.60 & 27.97 \\
    % PlenOctrees~\cite{yu2021plenoctrees} & 31.71 & 34.66 & 25.31 & 30.79 & 36.79 & 32.95 & 29.76 & 33.97 & 29.42 \\
    % Plenoxels~\cite{yu2021plenoxels} & 31.71 & 33.98 & 25.35 & 31.83 & 36.43 & 34.10 & 29.14 & 33.26 & 29.62 \\
    % DVGO~\cite{sun2021direct} & 31.95 & 34.09 & 25.44 & 32.78 & 36.74 & 34.64 & 29.57 & 33.20 & 29.13 \\

    % \hline
    % Ours-CP-384  & 31.56 & 33.60 & 25.17 & 30.72 & 36.24 & 34.05 & 30.10 & 33.77 & 28.84 \\
    % Ours-VM-192-SH  & 32.00 & 34.68 & 25.37 & 32.30 & 36.30 & 35.42 & 29.30 & 33.21 & 29.46 \\
    % Ours-VM-48  & 32.39 & 34.68 & 25.58 & 33.37 & 36.81 & 35.51 & 29.45 & 33.59 & 30.12 \\
    % Ours-VM-192-15k & 32.52 & 34.95 & 25.63 & 33.46 & 36.85 & 35.78 & 29.78 & 33.69 & 30.04\\
    % Ours-VM-192-30k & 33.14 & 35.76 & 26.01 & 33.99 & 37.41 & 36.46 & 30.12 & 34.61 & 30.77 \\
    % \hline
    
    % \hline
    
    % \hline    

    \multicolumn{9}{@{}l}{\rule{0pt}{3ex}\bf SSIM$\uparrow$} \\
    \hline
    SRN~\cite{sitzmann2019scene} & 0.846 & 0.910 & 0.766 & 0.849 & 0.923 & 0.809 & 0.808 & 0.947 & 0.757 \\
    NeRF~\cite{mildenhall2020nerf} & 0.947 & 0.967 & 0.925 & 0.964 & 0.974 & 0.961 & 0.949 & 0.980 & 0.856 \\
    NSVF~\cite{liu2020neural} & 0.953 & 0.968 & 0.931 & 0.973 & 0.980 & 0.960 & 0.973 & 0.987 & 0.854 \\
    SNeRG~\cite{hedman2021baking} & 0.950 & 0.975 & 0.929 & 0.967 & 0.971 & 0.973 & 0.938 & 0.982 & 0.865 \\
    PlenOctrees~\cite{yu2021plenoctrees} & 0.958 & 0.981 & 0.933 & 0.970 & \gold{0.982} & 0.971 & \gold{0.955} & 0.987 & 0.884 \\
    Plenoxels~\cite{yu2021plenoxels} & 0.958 & 0.977 & 0.933 & 0.976 & 0.980 & 0.976 & 0.949 & 0.985 & 0.890 \\
    DVGO~\cite{sun2021direct} & 0.957 & 0.977 & 0.930 & 0.978 & 0.980 & 0.976 & 0.951 & 0.983 & 0.879 \\
    \hline
    Ours-CP-384  & 0.949 & 0.973 & 0.921 & 0.965 & 0.975 & 0.971 & 0.950 & 0.983 & 0.857 \\
    Ours-VM-192-SH    & 0.955 & 0.979 & 0.928 & 0.976 & 0.977 & 0.978 & 0.941 & 0.983 & 0.875  \\
    Ours-VM-48  & 0.957 & 0.980 & 0.929 & 0.979 & 0.979 & 0.979 & 0.942 & 0.984 & 0.883 \\
    Ours-VM-192-15k & 0.959 & 0.982 & 0.933 & 0.981 & 0.980 & 0.981 & 0.949 & 0.985 & 0.886\\
    Ours-VM-192-30k &\gold{ 0.963} & \gold{0.985} & \gold{0.937} & \gold{0.982} & \gold{0.982} & \gold{0.983} & 0.952 & \gold{0.988} & \gold{0.895} \\
    \hline
    
    \hline
    
    \hline

    \multicolumn{9}{@{}l}{\rule{0pt}{3ex}\bf LPIPS$_{VGG}\downarrow$ } \\
    \hline
    SRN~\cite{sitzmann2019scene} & 0.170 & 0.106 & 0.267 & 0.149 & 0.100 & 0.200 & 0.174 & 0.063 & 0.299 \\
    NeRF~\cite{mildenhall2020nerf} & 0.081 & 0.046 & 0.091 & 0.044 & 0.121 & 0.050 & 0.063 & 0.028 & 0.206 \\

    PlenOctrees~\cite{yu2021plenoctrees}  & 0.053 & 0.022 & 0.076 & 0.038 & \gold{0.032} & 0.034 & 0.059 & 0.017 & 0.144 \\
    Plenoxels~\cite{yu2021plenoxels}  & 0.049 & 0.031 & \gold{0.067} & 0.026 & 0.037 & 0.028 & \gold{0.057} &\gold{0.015} & \gold{0.134} \\
    DVGO~\cite{sun2021direct} & 0.053 & 0.027 & 0.077 & 0.024 & 0.034 & 0.028 & 0.058 & 0.017 & 0.161 \\
    % \hline
    \hline
    Ours-CP-384       & 0.076 & 0.044 & 0.114 & 0.058 & 0.052 & 0.038 & 0.068 & 0.035 & 0.196 \\
    Ours-VM-192-SH    & 0.058 & 0.031 & 0.082 & 0.028 & 0.048 & 0.024 & 0.069 & 0.022 & 0.160 \\
    Ours-VM-48        & 0.057 & 0.030 & 0.087 & 0.028 & 0.039 & 0.024 & 0.072 & 0.021 & 0.155 \\
    Ours-VM-192-15k       & 0.053 & 0.026 & 0.078 & 0.025 & 0.038 & 0.021 & 0.063 & 0.020 & 0.153 \\
    Ours-VM-192-30k       & \gold{0.047} & \gold{0.022} & 0.073 & \gold{0.022} & \gold{0.032} & \gold{0.018} & 0.058 & \gold{0.015} & 0.138 \\

    %s
    \hline
    \multicolumn{9}{@{}l}{\rule{0pt}{3ex}\bf LPIPS$_{Alex}\downarrow$} \\
    NSVF~\cite{liu2020neural}  & 0.047 & 0.043 & 0.069 & 0.017 & 0.025 & 0.029 & 0.021 & 0.010 & 0.162 \\
    DVGO~\cite{sun2021direct}  & 0.035 & 0.016 & 0.061 & 0.015 & 0.017 & 0.014 & \gold{0.026} & 0.014 & 0.118 \\

     \hline
    Ours-CP-384        & 0.041 & 0.022 & 0.069 & 0.024 & 0.024 & 0.014 & 0.031 & 0.018 & 0.130 \\
    Ours-VM-192-SH     & 0.058 & 0.031 & 0.082 & 0.028 & 0.048 & 0.024 & 0.069 & 0.022 & 0.160  \\
    Ours-VM-48         & 0.032 & 0.014 & 0.059 & 0.015 & 0.017 & 0.009 & 0.036 & 0.012 & 0.098 \\
    Ours-VM-192-15k        & 0.032 & 0.013 & 0.056 & 0.014 & 0.017 & 0.009 & 0.029 & 0.013 & 0.101 \\
    Ours-VM-192-30k        & \gold{0.027} & \gold{0.010} & \gold{0.051} & \gold{0.012} & \gold{0.013} & \gold{0.007} & \gold{0.026} & \gold{0.009} & \gold{0.085} \\
    \hline

    \end{tabular}
    \caption{Quantitative results on each scene from the {\bf Synthetic-NeRF}~\cite{mildenhall2020nerf} dataset. %Ours $^\dag$ refers to our tensorf representation with direct optimization (i.e., without network) while ours are trained with a small MLP network.
    }
    \label{tab:supp_breakdown_nerf}
    \vspace{2em}
\end{table*}

\newpage
\begin{figure*}[htbp]
    \includegraphics[width=\linewidth]{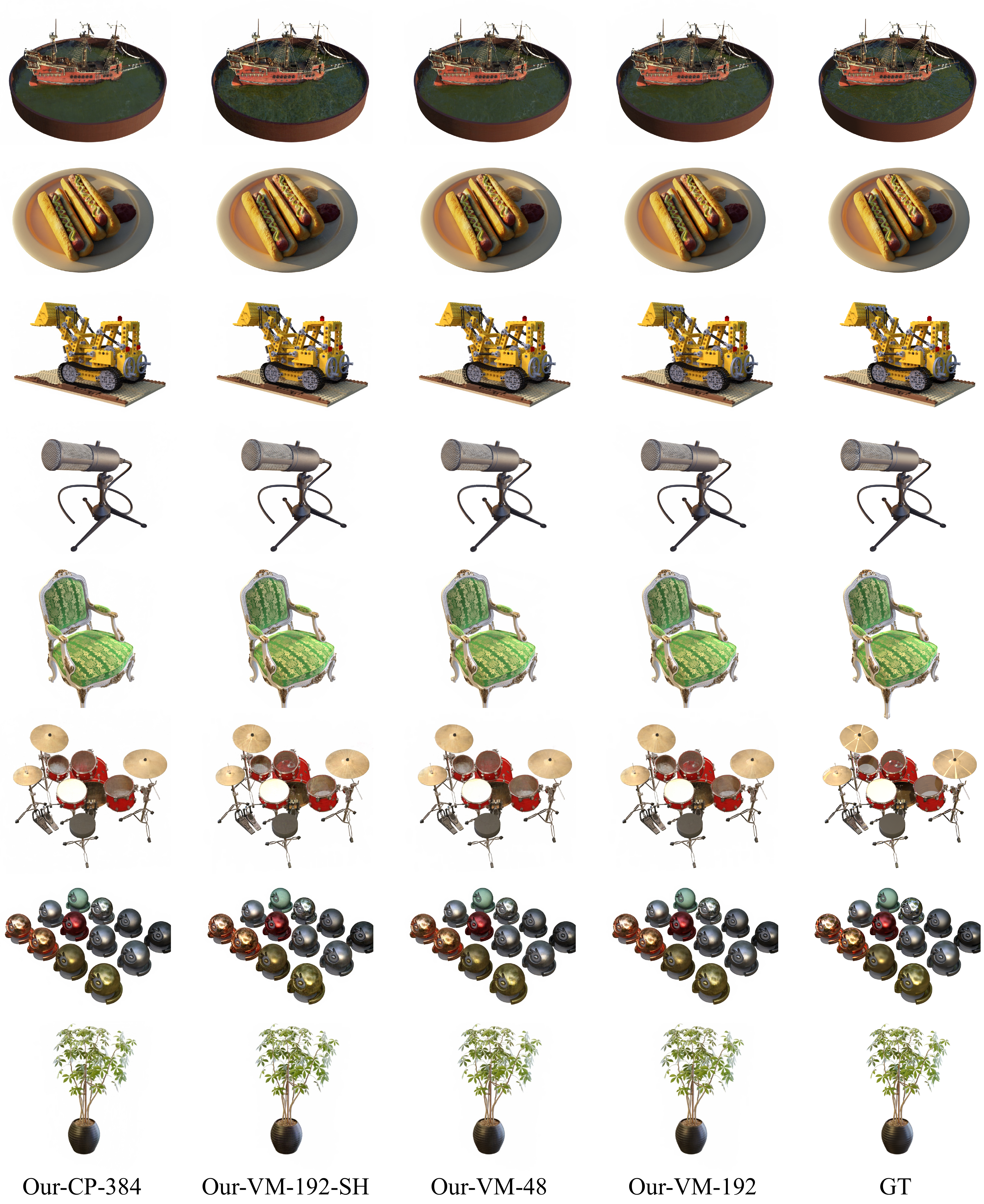}
    \caption{Our rendering results on {\bf Synthetic-NeRF} dataset. From top to bottom: Ship, Hotdog, Lego, Mic, Chair, Drums, Materials, Ficus.}
    \label{fig:synthetic_nerf}
\end{figure*}
%%%%%%%%%%%%%%%%%%%%%%%%%%%%%%%%%%%%%%%%%%%%%%%%%%%%%%%%%%%%%%%%%%%%%%%%%%%%%%%%%%%%%%%%%%%

\begin{table*}[htpb]
    \centering
    \begin{tabular}{l|c|cccccccc}
    \hline
    Methods & Avg. & {\it \tiny{Wineholder}} & {\it \tiny{Steamtrain}} & {\it \tiny{Toad}} & {\it \tiny{Robot}} & {\it \tiny{Bike}} & {\it \tiny{Palace}} & {\it \tiny{Spaceship}} & {\it \tiny{Lifestyle}} \\
    \hline\hline
    \multicolumn{9}{@{}l}{\rule{0pt}{3ex}\bf PSNR$\uparrow$} \\
    \hline
    SRN~\cite{sitzmann2019scene} & 24.33 & 20.74 & 25.49 & 25.36 & 22.27 & 23.76 & 24.45 & 27.99 & 24.58 \\
    NeRF~\cite{mildenhall2020nerf} & 30.81 & 28.23 & 30.84 & 29.42 & 28.69 & 31.77 & 31.76 & 34.66 & 31.08 \\
    NSVF~\cite{liu2020neural} & 35.13 & 32.04 & 35.13 & 33.25 & 35.24 & 37.75 & 34.05 & \gold{39.00} & \gold{34.60} \\
    DVGO~\cite{sun2021direct} &35.08 &30.26 &36.56 &33.10 &36.36 &38.33 &34.49 &37.71& 33.79\\
    \hline
    Ours-CP-384     & 34.48 & 29.92 & 36.07 & 31.37 & 35.92 & 36.74 & 36.26 & 37.01 & 32.54 \\
    Ours-VM-192-SH  & 35.30 & 29.72 & 37.33 & 34.03 & 37.59 & 38.61 & 36.09 & 35.82 & 33.21  \\
    Ours-VM-48      & 35.34 & 30.46 & 37.06 & 33.13 & 36.92 & 37.98 & 36.32 & 37.19 & 33.68 \\
    Ours-VM-192-15k & 35.59 & 30.31 & 37.20 & 33.63 & 37.29 & 38.33 & 36.57 & 37.77 & 33.62\\
    Ours-VM-192-30k & \gold{36.52} & \gold{31.32} & \gold{37.87} & \gold{34.85} & \gold{38.26} & \gold{39.23} & \gold{37.56} & 38.60 & 34.51 \\
    \hline
    
    \hline
    
    \hline    
    
    \multicolumn{9}{@{}l}{\rule{0pt}{3ex}\bf SSIM$\uparrow$} \\
    \hline
    SRN~\cite{sitzmann2019scene} & 0.882 & 0.850 & 0.923 & 0.822 & 0.904 & 0.926 & 0.792 & 0.945 & 0.892 \\
    NeRF~\cite{mildenhall2020nerf} & 0.952 & 0.920 & 0.966 & 0.920 & 0.960 & 0.970 & 0.950 & 0.980 & 0.946 \\
    NSVF~\cite{liu2020neural} & 0.979 & \gold{0.965} & 0.986 & 0.968 & 0.988 & 0.991 & 0.969 & \gold{0.991} & \gold{0.971} \\
    DVGO~\cite{sun2021direct} &0.975& 0.949& 0.989& 0.966 &0.992& 0.991& 0.962& 0.988& 0.965 \\
    \hline
    Ours-CP-384     & 0.971 & 0.947 & 0.986 & 0.950 & 0.990 & 0.987 & 0.971 & 0.984 & 0.951  \\
    Ours-VM-192-SH  & 0.977 & 0.953 & 0.988 & 0.974 & 0.993 & 0.991 & 0.972 & 0.982 & 0.964  \\
    Ours-VM-48      & 0.976 & 0.952 & 0.988 & 0.968 & 0.992 & 0.990 & 0.973 & 0.985 & 0.962 \\
    Ours-VM-192-15k & 0.978 & 0.953 & 0.989 & 0.972 & 0.993 & 0.991 & 0.975 & 0.987 & 0.964 \\
    Ours-VM-192-30k & \gold{0.982} & 0.961 & \gold{0.991} & \gold{0.978} & \gold{0.994} & \gold{0.993} & \gold{0.979} & 0.989 & 0.968 \\
    \hline

    % \multicolumn{9}{@{}l}{\rule{0pt}{3ex}\bf LPIPS$\downarrow$ {\footnotesize (Vgg)}} \\
    \hline
    
    \hline

    \multicolumn{9}{@{}l}{\rule{0pt}{3ex}\bf LPIPS$_{VGG}\downarrow$} \\
    \hline
    DVGO~\cite{sun2021direct} & 0.033 & 0.055 & 0.019 & 0.047 & 0.013 & 0.011 & 0.043 & \gold{0.019} & 0.054\\
    Ours-CP-384             & 0.045 & 0.082 & 0.031 & 0.067 & 0.016 & 0.023 & 0.031 & 0.028 & 0.084 \\
    Ours-VM-192-SH          & 0.031 & 0.057 & 0.024 & 0.035 & 0.011 & 0.013 & 0.030 & 0.026 & 0.051  \\
    Ours-VM-48              & 0.034 & 0.061 & 0.023 & 0.047 & 0.013 & 0.014 & 0.029 & 0.025 & 0.059 \\
    Ours-VM-192-15k         & 0.031 & 0.060 & 0.020 & 0.040 & 0.011 & 0.012 & 0.028 & 0.022 & 0.055\\
    Ours-VM-192-30k         & \gold{0.026} & \gold{0.051} & \gold{0.017} & \gold{0.031} & \gold{0.010} & \gold{0.010} & \gold{0.022} & 0.020 & \gold{0.048} \\
    
    \hline

    \multicolumn{9}{@{}l}{\rule{0pt}{3ex}\bf LPIPS$_{Alex}\downarrow$} \\
    \hline
    SRN~\cite{sitzmann2019scene} & 0.141 & 0.224 & 0.082 & 0.204 & 0.120 & 0.075 & 0.240 & 0.061 & 0.120 \\
    NeRF~\cite{mildenhall2020nerf} & 0.043 & 0.096 & 0.031 & 0.069 & 0.038 & 0.019 & 0.031 & 0.016 & 0.047 \\
    NSVF~\cite{liu2020neural} & 0.015 & \gold{0.020} & 0.010 & 0.032 & 0.007 & 0.004 & 0.018 & \gold{0.006} & \gold{0.020} \\
    DVGO~\cite{sun2021direct} & 0.019 &0.038 &0.010 &0.030& 0.005 &0.004 &0.027 &0.009 &0.027\\
    \hline
    Ours-CP-384  & 0.021 & 0.040 & 0.010 & 0.039 & 0.006 & 0.007 & 0.014 & 0.015 & 0.042  \\
    Ours-VM-192-SH            & 0.015 & 0.030 & 0.008 & 0.021 & \gold{0.003} & \gold{0.003} & 0.016 & 0.016 & 0.025  \\
    Ours-VM-48         & 0.016 & 0.031 & 0.008 & 0.025 & 0.004 & 0.004 & 0.015 & 0.013 & 0.026 \\
    Ours-VM-192-15k        & 0.015 & 0.033 & 0.008 & 0.022 & 0.004 & 0.004 & 0.015 & 0.011 & 0.026 \\
    Ours-VM-192-30k        & \gold{0.012} & 0.024 & \gold{0.006} & 0.016 & \gold{0.003} & \gold{0.003} & \gold{0.011} & 0.009 & 0.021 \\

    \hline

    \end{tabular}
    \caption{Quantitative results on each scene from the {\bf Synthetic-NSVF}~\cite{liu2020neural} dataset. }
    \label{tab:supp_breakdown_nsvf}
\end{table*}

\begin{figure}[t]
    \includegraphics[width=\linewidth]{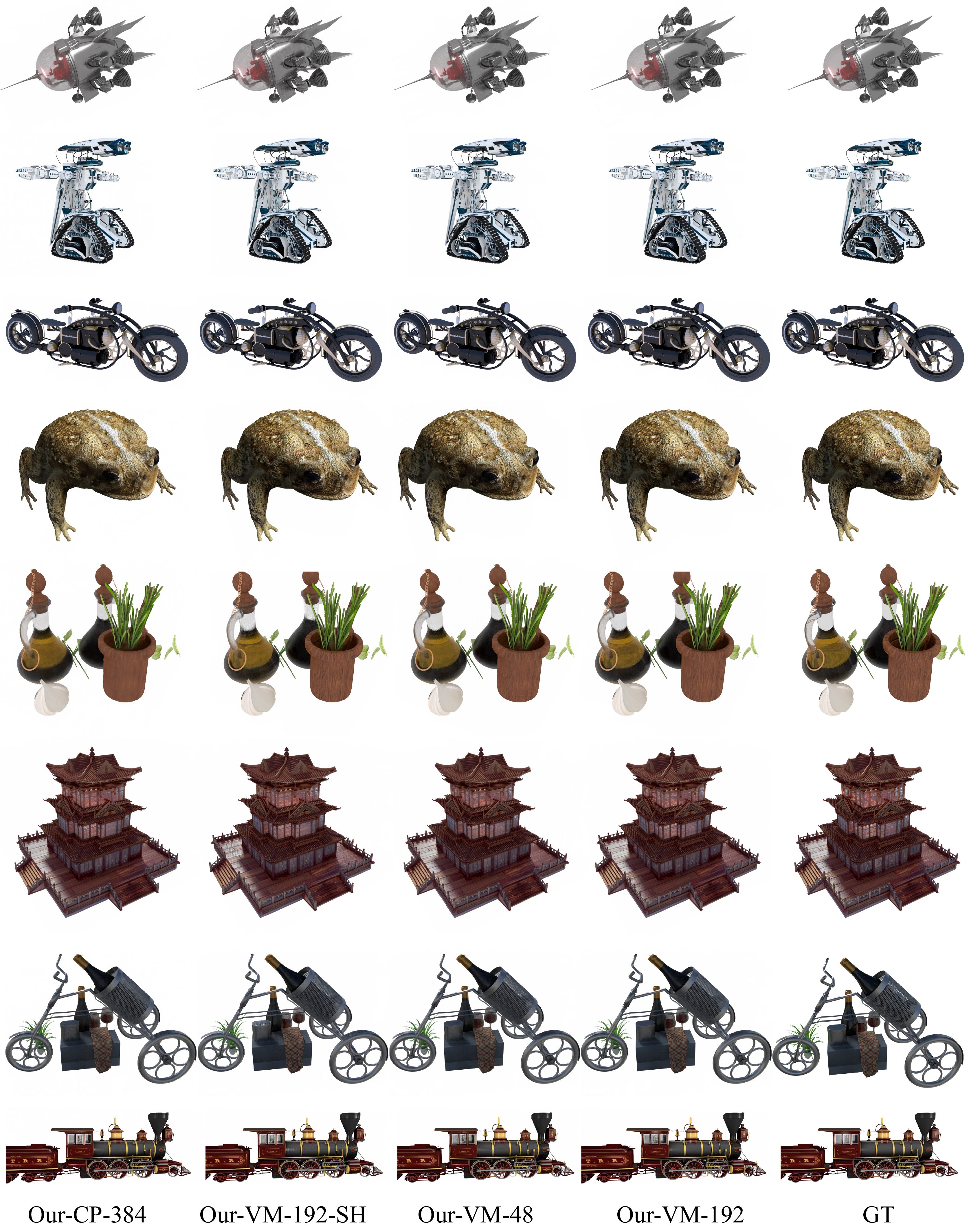}
    \caption{Our rendering results on {\bf NSVF} \cite{liu2020neural} dataset. From top to bottom: Spaceship, Robot, Toad, Lifestyle, Palace, Wineholder, Steamtrain.}
    \label{fig:synthetic_nsvf}
\end{figure}
%%%%%%%%%%%%%%%%%%%%%%%%%%%%%%%%%%%%%%%%%%%%%%%%%%%%%%%%%%%%%%%%%%%%%%%%%%%%%%%%%%%%

\begin{table*}[t]
    \centering
    \renewcommand\tabcolsep{5.0pt}
    \begin{tabular}{l|c|ccccc}
    \hline
    Methods & Avg. & {\it Ignatius} & {\it Truck} & {\it Barn} & {\it Caterpillar} & {\it Family} \\
    \hline\hline
    \multicolumn{7}{@{}l}{\rule{0pt}{3ex}\bf PSNR$\uparrow$} \\
    \hline
    SRN~\cite{sitzmann2019scene} & 24.10 & 26.70 & 22.62 & 22.44 & 21.14 & 27.57 \\
    NeRF~\cite{mildenhall2020nerf} & 25.78 & 25.43 & 25.36 & 24.05 & 23.75 & 30.29 \\
    NSVF~\cite{liu2020neural} & 28.48 & 27.91 & 26.92 & 27.16 & \gold{26.44} & 33.58 \\
    PlenOctrees~\cite{yu2021plenoctrees} & 27.99 & 28.19 & 26.83 & 26.80 & 25.29 & 32.85 \\
    Plenoxels~\cite{yu2021plenoxels} & 27.43 & 27.51 & 26.59 & 26.07 & 24.64 & 32.33  \\
    DVG ~\cite{sun2021direct} & 28.41 & 28.16 & \gold{27.15} & 27.01 & 26.00 & 33.75 \\
    \hline
    Ours-CP-384 & 27.59 & 27.86 & 26.25 & 26.74 & 24.73 & 32.39  \\
    Ours-VM-192-SH     & 27.81 & 27.78 & 26.73 & 26.03 & 25.37 & 33.12  \\
    Ours-VM-48  & 28.06 & 28.22 & 26.81 & 26.70 & 25.43 & 33.12 \\
    Ours-VM-192-15k & 28.07 & 28.27 & 26.57 & 26.93 & 25.35 & 33.22 \\
    Ours-VM-192-30k & \gold{28.56} & \gold{28.34} & 27.14 & \gold{27.22} & 26.19 & \gold{33.92} \\
    \hline
    
    \hline
    
    \hline    

    \multicolumn{7}{@{}l}{\rule{0pt}{3ex}\bf SSIM$\uparrow$} \\
    \hline
    SRN~\cite{sitzmann2019scene} & 0.847 & 0.920 & 0.832 & 0.741 & 0.834 & 0.908 \\
    NeRF~\cite{mildenhall2020nerf} & 0.864 & 0.920 & 0.860 & 0.750 & 0.860 & 0.932 \\
    NSVF~\cite{liu2020neural} & 0.901 & 0.930 & 0.895 & 0.823 & 0.900 & 0.954 \\
    PlenOctrees~\cite{yu2021plenoctrees} & 0.917 & \gold{0.948} & \gold{0.914} & 0.856 & 0.907 & 0.962 \\
    Plenoxels~\cite{yu2021plenoxels} & 0.906 & 0.943 & 0.901 & 0.829 & 0.902 & 0.956 \\
    DVGO~\cite{sun2021direct} &0.911 & 0.944 & 0.906 & 0.838 & 0.906 & 0.962\\
    \hline
    Ours-CP-384 & 0.897 & 0.934 & 0.885 & 0.839 & 0.879 & 0.948  \\
    Ours-VM-192-SH     & 0.907 & 0.942 & 0.900 & 0.834 & 0.897 & 0.960  \\
    Ours-VM-48  & 0.909 & 0.943 & 0.902 & 0.845 & 0.899 & 0.957 \\
    Ours-VM-192-15k & 0.913 & 0.944 & 0.905 & 0.855 & 0.902 & 0.960 \\
    Ours-VM-192-30k & \gold{0.920} & \gold{0.948} & \gold{0.914} & \gold{0.864} & \gold{0.912} & \gold{0.965} \\
    \hline
    
    \hline
    
    \hline    

    \multicolumn{7}{@{}l}{\rule{0pt}{3ex}\bf LPIP$_{VGG} \downarrow$} \\
    \hline
    PlenOctrees~\cite{yu2021plenoctrees} & 0.131 & 0.080 & 0.130 & 0.226 & \gold{0.148} & 0.069 \\
    Plenoxels~\cite{yu2021plenoxels} & 0.162 & 0.102 & 0.163 & 0.303 & 0.166 & 0.078 \\
    DVGO~\cite{sun2021direct} & 0.155 & 0.083 & 0.160 & 0.294 & 0.167 & 0.070 \\
    \hline
    Ours-CP-384     & 0.181 & 0.106 & 0.202 & 0.283 & 0.227 & 0.088 \\
    Ours-VM-192-SH  & 0.156 & 0.089 & 0.161 & 0.286 & 0.175 & 0.069  \\
    Ours-VM-48      & 0.155 & 0.085 & 0.161 & 0.278 & 0.177 & 0.074 \\
    Ours-VM-192-15k & 0.152 & 0.084 & 0.162 & 0.269 & 0.173 & 0.071 \\
    Ours-VM-192-30k & \gold{0.140} & \gold{0.078} & \gold{0.145} &\gold{0.252} & 0.159 & \gold{0.064} \\
    \hline

    \multicolumn{7}{@{}l}{\rule{0pt}{3ex}\bf LPIPS$_{Alex}\downarrow$} \\
    \hline
    SRN~\cite{sitzmann2019scene} & 0.251 & 0.128 & 0.266 & 0.448 & 0.278 & 0.134 \\ 
    NeRF~\cite{mildenhall2020nerf} & 0.198 & 0.111 & 0.192 & 0.395 & 0.196 & 0.098 \\
    NSVF~\cite{liu2020neural} & 0.155 & 0.106 & 0.148 & 0.307 & 0.141 & 0.063 \\
    DVGO~\cite{sun2021direct} & 0.148 & 0.090 & 0.145 & 0.290 & 0.152 & 0.064 \\
    \hline
    Ours-CP-384     & 0.144 & 0.089 & 0.154 & 0.237 & 0.176 & 0.063  \\
    Ours-VM-192-SH  & 0.164 & 0.098 & 0.168 & 0.309 & 0.175 & 0.072  \\
    Ours-VM-48      & 0.145 & 0.089 & 0.145 & 0.266 & 0.161 & 0.066 \\
    Ours-VM-192-15k & 0.140 & 0.087 & 0.150 & 0.240 & 0.157 & 0.066 \\
    Ours-VM-192-30k & \gold{0.125} & \gold{0.081} & \gold{0.129} & \gold{0.217} & \gold{0.139} & \gold{0.057} \\
    \hline

    \end{tabular}
    \caption{Quantitative results on each scene from the {\bf Tanks\&Temples} \cite{Knapitsch2017} dataset.}
    \label{tab:supp_breakdown_tanksandtemples}
\end{table*}

\begin{figure}[t]
    \includegraphics[width=\linewidth]{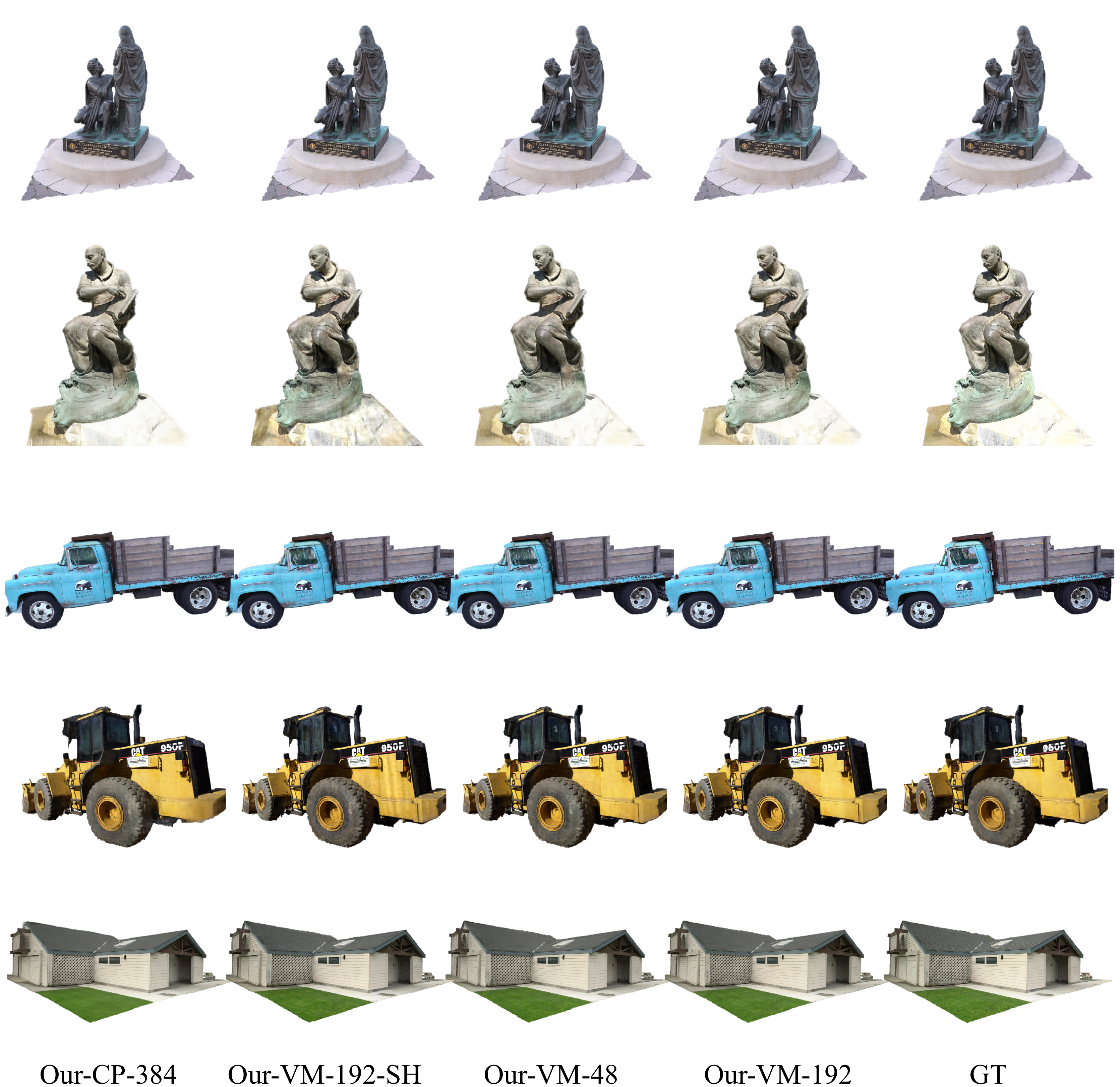}
    \caption{Our rendering results on {\bf Tanks\&Temples}\cite{Knapitsch2017} dataset. From top to bottom: Family, Ignatius, Truck, Caterpillar, Barn.}
    \label{fig:res_tt}
\end{figure}
%%%%%%%%%%%%%%%%%%%%%%%%%%%%%%%%%%%%%%%%%%%%%%%%%%%%%%%%%%%%%%%%%%%%%%%%%%%%%%%%%%%%%%%%%%%%%%%%

\begin{table*}[htpb]
    \centering
    \renewcommand\tabcolsep{2.0pt}
    \begin{tabular}{l|c|cccccccc}
    \hline
    Methods & Avg. & {\it Room} & {\it Fern} & {\it Leaves} & {\it Fortress} & {\it Orchids} & {\it Flower} & {\it T-Rex} & {\it Horns} \\
    \hline\hline
    \multicolumn{9}{@{}l}{\rule{0pt}{3ex}\bf PSNR$\uparrow$} \\
    \hline
    % SRN~\cite{sitzmann2019scene} & 24.33 & 20.74 & 25.49 & 25.36 & 22.27 & 23.76 & 24.45 & 27.99 & 24.58 \\
    NeRF~\cite{mildenhall2020nerf}   & 26.50 & \gold{32.70} &25.17 &20.92 &31.16 &\gold{20.36} & 27.40 &26.80 &27.45 \\
    Plenoxels~\cite{yu2021plenoxels} & 26.29 & 30.22 & 25.46 & \gold{21.41} & 31.09 & 20.24 & 27.83 & 26.48 & 27.58 \\
    % DVGO~\cite{sun2021direct} & \silver{35.08} & \silver{30.26} & \gold{36.56} & \silver{33.10} & \gold{36.36} & \gold{38.33} & \gold{34.49} & \silver{37.71} & \silver{33.79} \\
    \hline
    % CP-Neural-192 & 32.31 & 35.36 & 25.38 & 32.12& \silver{36.25} & 35.65 & 29.28 & 34.63 & 29.85  \\
    % Our-VM-96-SH     & 0.955 & 0.941 & 0.875 & 0.978 & 0.983 & 0.928 & 0.979 & 0.976 & 0.977  \\
    Ours-VM-48 & 26.51 & 31.80 & 25.31 & 21.34 & 31.14 & 20.02 & 28.22 & 26.61 & 27.64\\
    Our-VM-96 & \gold{26.73} & 32.35 & \gold{25.27} & 21.30 & \gold{31.36} & 19.87 & \gold{28.60} & \gold{26.97} & \gold{28.14} \\
    \hline
    
    \hline
    
    \hline    
    
    \multicolumn{9}{@{}l}{\rule{0pt}{3ex}\bf SSIM$\uparrow$} \\
    \hline
    % SRN~\cite{sitzmann2019scene} & 0.882 & 0.850 & 0.923 & 0.822 & 0.904 & 0.926 & 0.792 & 0.945 & 0.892 \\
    NeRF~\cite{mildenhall2020nerf}   & 0.811 & 0.948 & 0.792 & 0.690 & 0.881 & 0.641 & 0.827 & 0.880 & 0.828\\
    Plenoxels~\cite{yu2021plenoxels} & \gold{0.839} & 0.937 & \gold{0.832} & \gold{0.760} & 0.885 & \gold{0.687} & 0.862 & 0.890 & 0.857\\

    \hline
    % CP-Neural-192 & 32.31 & 35.36 & 25.38 & 32.12& \silver{36.25} & 35.65 & 29.28 & 34.63 & 29.85  \\
    % Our-VM-96-SH     & 0.955 & 0.941 & 0.875 & 0.978 & 0.983 & 0.928 & 0.979 & 0.976 & 0.977  \\
    Ours-VM-48 & 0.832 & 0.946 & 0.816 & 0.746 & 0.889 & 0.655 & 0.859 & 0.890 & 0.859 \\
    Ours-VM-96 & \gold{0.839} & \gold{0.952} & 0.814 & 0.752 & \gold{0.897} & 0.649 & \gold{0.871} & \gold{0.900} & \gold{0.877} \\
    \hline

    % \multicolumn{9}{@{}l}{\rule{0pt}{3ex}\bf LPIPS$\downarrow$ {\footnotesize (Vgg)}} \\
    \hline

    \hline

    \multicolumn{9}{@{}l}{\rule{0pt}{3ex}\bf LPIPS$_{VGG}\downarrow$ } \\
    \hline
    % SRN~\cite{sitzmann2019scene} & 0.141 & 0.224 & 0.082 & 0.204 & 0.120 & 0.075 & 0.240 & 0.061 & 0.120 \\
    NeRF~\cite{mildenhall2020nerf}   & 0.250 & 0.178 &0.280 &0.316 &0.171   &0.321 & 0.219 & 0.249 & 0.268 \\
    Plenoxels~\cite{yu2021plenoxels} & 0.210 & 0.192 & \gold{0.224} & \gold{0.198} & 0.180 & \gold{0.242} & 0.179 & 0.238 & 0.231 \\

    \hline
    % CP-Neural-192  & 32.31 & 35.36 & 25.38 & 32.12& \silver{36.25} & 35.65 & 29.28 & 34.63 & 29.85  \\
    % Our-VM-96-SH      & 0.058 & 0.069 & 0.160 & 0.024 & 0.022 & 0.082 & 0.031 & 0.028 & 0.048  \\
    Ours-VM-48 & 0.217 & 0.181 & 0.237 & 0.230 & 0.159 & 0.283 & 0.187 & 0.236 & 0.221 \\
    Ours-VM-96 & \gold{0.204} & \gold{0.167} & 0.237 & 0.217 & \gold{0.148} & 0.278 & \gold{0.169} & \gold{0.221} & \gold{0.196} \\
    \hline

    \multicolumn{9}{@{}l}{\rule{0pt}{3ex}\bf LPIPS$_{Alex} \downarrow$ } \\
    \hline
    % CP-Neural-192 (Alex) & 32.31 & 35.36 & 25.38 & 32.12& \silver{36.25} & 35.65 & 29.28 & 34.63 & 29.85  \\
    % Our-VM-96-SH     & 0.015 & 0.030 & 0.008 & 0.021 & 0.003 & 0.003 & 0.016 & 0.016 & 0.025  \\
    Ours-VM-48 & 0.135 & 0.093 & 0.161 & 0.167 & 0.084 & 0.204 & 0.121 & 0.108 & 0.146 \\
    Ours-VM-96 & 0.124 & 0.082 & 0.155 & 0.153 & 0.075 & 0.201 & 0.106 & 0.099 & 0.123 \\
    \hline
    
    \end{tabular}
    \caption{Quantitative results on each scene from the {\bf forward-facing}~\cite{liu2020neural} dataset. }
    \label{tab:supp_breakdown_llff}
\end{table*}

\begin{figure}[t]
    \includegraphics[width=\linewidth]{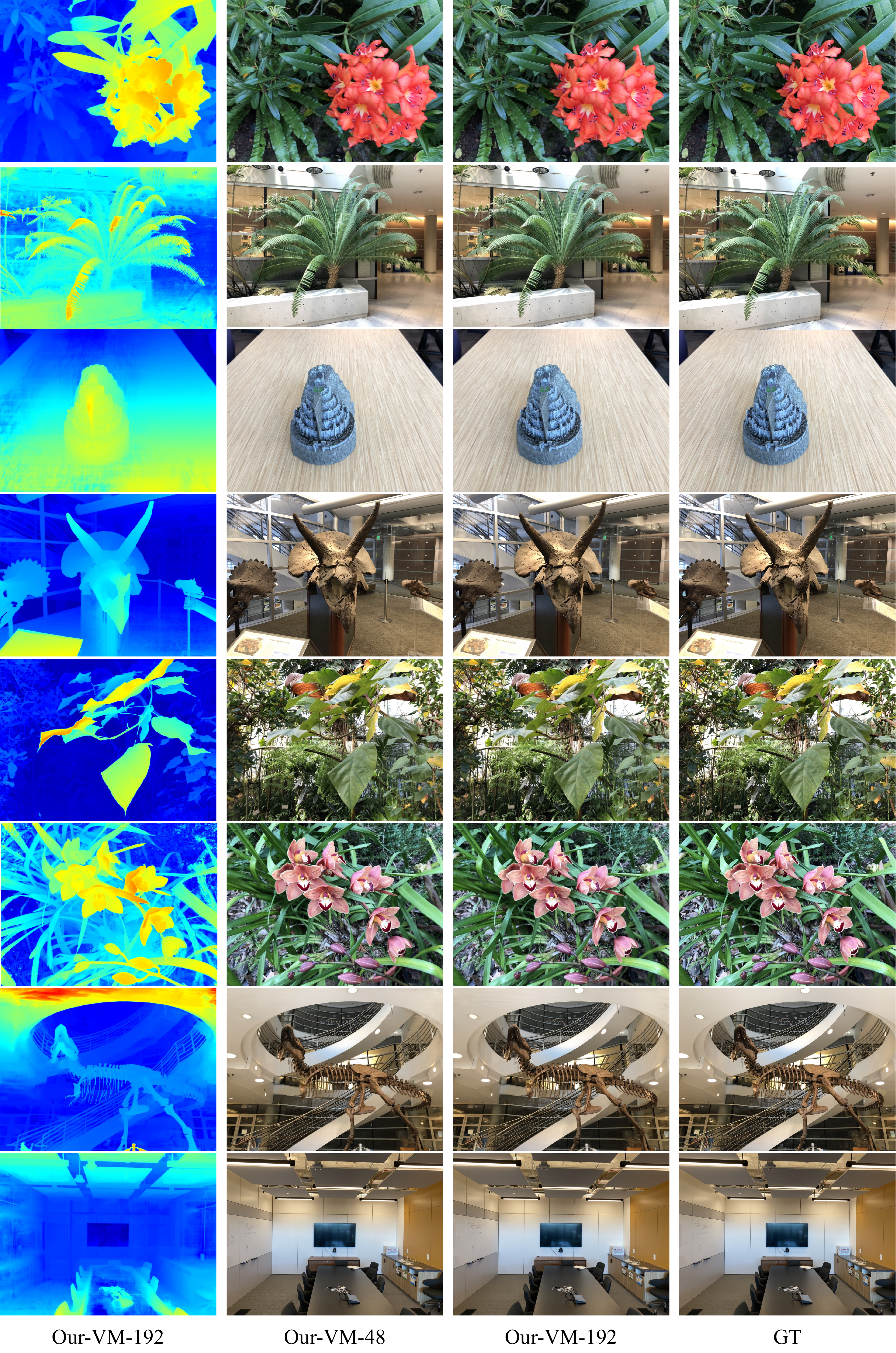}
    \caption{Our rendering results on {\bf forward-facing}~\cite{liu2020neural} dataset. From top to bottom: Flower, Fern, Fortress, Horn, Leaves, Orchids, T-Rex, Room. }
    \label{fig:res_llff}
\end{figure}
\end{document}